\documentclass{article}

 \usepackage[preprint]{neurips_2026}

\usepackage[utf8]{inputenc} % allow utf-8 input
\usepackage[T1]{fontenc}    % use 8-bit T1 fonts
\usepackage{hyperref}       % hyperlinks
\usepackage{url}            % simple URL typesetting
\usepackage{booktabs}       % professional-quality tables
\usepackage{amsfonts}       % blackboard math symbols
\usepackage{nicefrac}       % compact symbols for 1/2, etc.
\usepackage{microtype}      % microtypography
\usepackage{xcolor}         % colors
\usepackage{graphicx}
\usepackage{wrapfig}
\usepackage{amsmath}
\usepackage{algorithm}
\usepackage{algorithmic}
\usepackage{makecell}
\usepackage{subcaption}
\usepackage{enumitem}
\usepackage{listings}
\usepackage{fontawesome5}
\newcommand{\method}{\textsc{Resd}}

\title{Learning with Rare Success but Rich Feedback via Reflection-Enhanced Self-Distillation}

\author{%
  Yuwei Zhang$^1$\thanks{Work done during internship.}\quad Sha Li$^2$\quad Changlong Yu$^2$\quad Qin Lu$^2$\quad Shuowei Jin$^2$\quad Chengyu Dong$^2$\\
  \textbf{Haoran Liu}$^1$\quad \textbf{Ilgee Hong}$^3$\footnotemark[1]\quad \textbf{Xintong Li}$^1$\footnotemark[1]\quad \textbf{Zhenyu Shi}$^2$\quad \textbf{Bing Yin}$^2$\quad \textbf{Jingbo Shang}$^1$\thanks{Corresponding authors.}\\
  UC San Diego$^1$\quad Amazon$^2$\quad Georgia Institute of Technology$^3$\\
  \texttt{\{yuz163,jshang\}@ucsd.edu}\\
}

\begin{document}

\maketitle

\begin{center}
    \href{https://yuweizhang.notion.site/resd}{\textcolor{blue}{\faGlobe~Project Page}} \quad
    \href{https://github.com/horizon-llm/RESD}{\textcolor{blue}{\faGithub~GitHub}}
\end{center}

\begin{abstract}
    % Self-distillation has emerged as a paradigm wherein a single language model iteratively refines its own capabilities by leveraging both reward-verified trajectories and environment feedback. We find, however, that reasoning models heavily rely on successful demonstrations, and struggle to translate failures into policy improvements when success is sparse. To bridge this gap, we introduce Reflection-Enhanced Self-Distillation (\method{}), a framework that converts failed trajectories into retrospective feedback, enabling the teacher model to synthesize corrective supervision even in the absence of successful rollouts.
    Enabling Large Language Models (LLMs) to continuously improve from environmental interactions is a central challenge in post-training. While on-policy self-distillation offers a promising paradigm, existing methods predominantly treat environmental feedback as a passive conditioning signal. Consequently, they heavily rely on successful demonstrations and struggle to learn in rare-success regimes. To bridge this gap, we introduce Reflection-Enhanced Self-Distillation (\method{}), a framework that transforms raw failure feedback into an active source of corrective supervision. Instead of passively appending feedback, \method{} interprets failed trajectories by generating retrospective reflections to diagnose local errors, and curates a persistent global playbook to preserve reusable lessons across training steps. The enriched context enables the self-teacher to provide actionable token-level supervision even in the absence of successful rollouts. Empirical evaluations on multiple continual learning tasks demonstrate that \method{} substantially outperforms standard self-distillation baselines. Furthermore, \method{} achieves significantly faster early-stage improvement than GRPO with $8\times$ samples using only a single rollout per prompt, highlighting its superior interaction efficiency.
    % On-policy self-distillation has emerged as a promising paradigm for improving Large Language Models (LLMs) under practical post-training budgets. By instantiating a self-teacher conditioned on privileged feedback, models can derive dense, token-level supervision directly from their own trajectories. However, existing self-distillation methods rely heavily on the availability of successful peer demonstrations, suffering severe degradation in rare-success regimes. We identify feedback formulation as a critical bottleneck in these settings; exposing the teacher to raw, unstructured failure signals yields weak supervisions.
    % To address this, we propose Reflection-Enhanced Self-Distillation (\method{}), a framework that actively interprets and structures environment feedback. For each failed rollout, \method{} generates a retrospective reflection to diagnose the specific error and curates recurring lessons into a persistent playbook. This enriches the self-teacher's context, transforming raw failure signals into actionable and reusable corrective supervision.
    % Empirical evaluations demonstrate that RESD substantially outperforms standard self-distillation baselines, particularly in rare-success regimes. Furthermore, RESD achieves faster early-stage improvement than reward-based methods like GRPO despite using 8x fewer environment interactions per prompt. These results demonstrate high sample efficiency and highlight the critical role of structured feedback in autonomous model self-improvement.
\end{abstract}

\section{Introduction}
\label{sec:intro}

\begin{wrapfigure}{r}{0.41\textwidth}
    \centering
    \vspace{-1em}
    \includegraphics[width=0.40\textwidth]{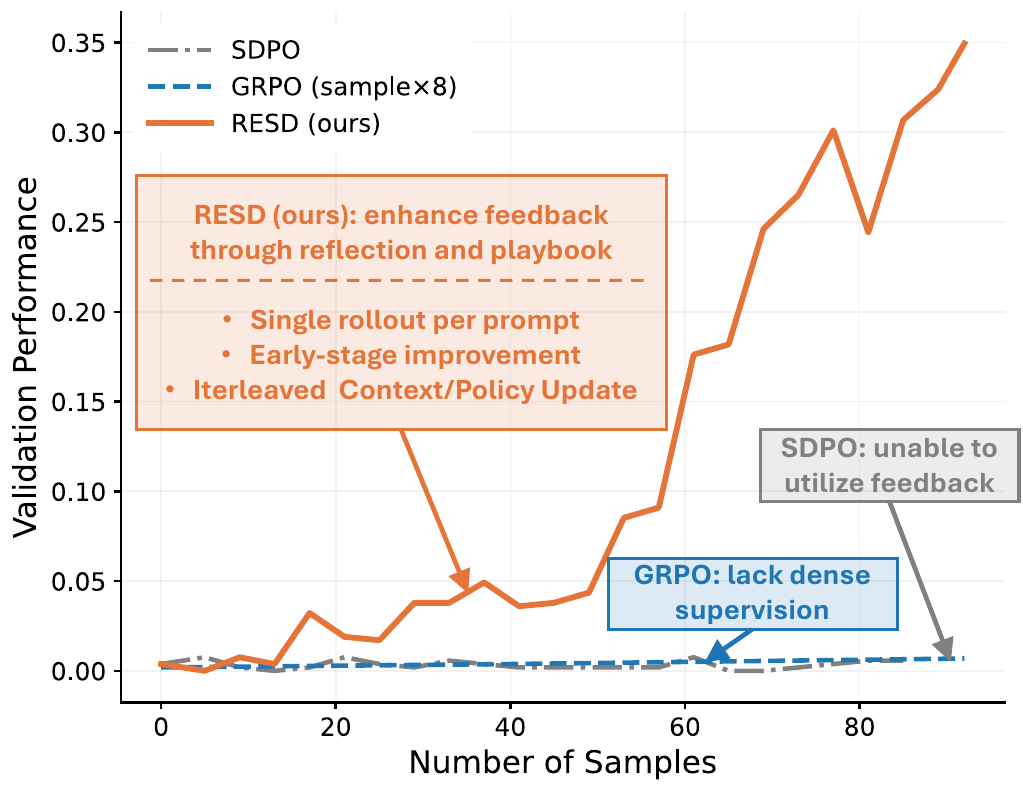}
    \caption{\method{} improves interaction efficiency during training. The x-axis is the number of samples.}
    \label{fig:intro}
    \vspace{-1.5em}
\end{wrapfigure}

A fundamental challenge in the post-training of Large Language Models (LLMs) is enabling continuous improvement through environmental interactions. Traditionally, Reinforcement Learning (RL) algorithms, such as PPO or GRPO, have been the standard paradigm for aligning models with desired outcomes. However, a critical bottleneck in deploying RL for complex, multi-step tasks is the reliance on sparse reward signals. In scenarios where successful trajectories are exceedingly rare, standard RL often struggles to effectively bootstrap policies because it lacks dense supervision to guide the model through a vast exploration space.

To mitigate the sparse reward problem, On-Policy Distillation (OPD) has emerged as a promising alternative~\citep{agarwal2023gkd,lu2025onpolicydistillation,xiao2026mimo}. Rather than relying on a single scalar reward at the end of a rollout, OPD leverages a stronger or privileged teacher model to compute target probabilities for every single token in the generated trajectory. This mechanism elegantly translates sparse trajectory-level outcomes into dense, token-level learning signals. However, standard OPD requires maintaining a separate expert model, which incurs high computational costs and risks distribution mismatch between the teacher and the student. To address this, self-distillation variants like SDPO instantiate the teacher from the model itself~\citep{hubotter2026reinforcement, zhao2026self}. By conditioning the self-teacher on environmental feedback, SDPO eliminates the need for an external oracle while perfectly aligning the supervisory signal with the model's own generation distribution.

While self-distillation resolves both reward sparsity and distribution mismatch, its success hinges entirely on the quality of the self-teacher's supervision. Current methods predominantly treat environmental feedback as a static, passive conditioning variable. \emph{We argue that this overlooks a key design choice: the structural representation of the feedback itself.} In sparse-reward regimes where successful demonstrations are absent, the teacher is forced to derive supervision almost entirely from failed rollouts. We demonstrate in Figure~\ref{fig:intro} that simply feeding raw failure outcomes or unstructured environment errors into the teacher context yields only marginal student improvements.
% The teacher, constrained by the same baseline capabilities as the student, cannot synthesize corrective token-level distributions from uninterpreted failures.

%shuowei finishes here
% \textcolor{blue}{version 2 ends}

Motivated by this observation, we propose Reflection-Enhanced Self-Distillation (\method{}), a framework that enriches the teacher context with actively interpreted feedback. For each failed rollout, \method{} first generates a retrospective reflection that identifies the likely cause of failure and the correction that would have avoided it. It then organizes recurring lessons into a persistent playbook, allowing future teacher prompts to reuse feedback-derived knowledge across training steps. The resulting enriched teacher context transforms raw failure feedback from a passive diagnostic signal into an actionable and reusable source of corrective supervision, enabling self-distillation to improve even when successful demonstrations are unavailable.

Empirically, we evaluate \method{} on tasks designed to reflect continuous learning beyond common post-training distributions. These tasks are novel to the model, often begin in rare-success regimes, and provide rich execution feedback despite receiving sparse binary rewards. Across these settings, \method{} substantially improves over standard self-distillation baselines and achieves faster early-stage improvement than GRPO while using only a single rollout per prompt. These results suggest that structured feedback enables sample-efficient bootstrapping from failures, while reward-based optimization remains a complementary tool once sufficient successful rollouts are available.
Our contributions are as follows:
\begin{itemize}[leftmargin=*]
    \item We identify \emph{feedback formulation} as a critical design axis for on-policy self-distillation especially in sparse-reward regimes.
    \item We propose \emph{Reflection-Enhanced Self-Distillation} (\method{}), which enriches the teacher context with retrospective reflections and a persistent playbook. By interpreting failures and preserving recurring lessons across training steps, \method{} turns raw feedback from a passive diagnostic signal into reusable corrective supervision. Empirically, we characterize the sample-efficiency of feedback-enhanced self-distillation under an online streaming protocol.
    \item More broadly, by revealing that self-improving models are highly sensitive to how feedback is structured, we position our reflection and playbook mechanisms as a plug-and-play module that can readily enhance other self-distillation objectives.
\end{itemize}

\section{Preliminaries}\label{sec:prelim}

Recent advancements such as Self-Distillation Policy Optimization (SDPO)~\citep{hubotter2026reinforcement} and On-Policy Self-Distillation (OPSD)~\citep{zhao2026self} introduce a ``self-teacher'' paradigm that replaces sparse outcomes in Reinforcement Learning with dense, token-level supervisions without the need for a massive separate teacher model. In these frameworks, a ``self-teacher'' policy is typically instantiated as the model's current or moving-average weights $\theta_{\text{old}}$, operating in a privileged context. The teacher is conditioned on both the intermediate state $s_t=(x_t, y_{<t})$ and the retrospective feedback $c(x,y)$, producing a corrected, feedback-informed distribution $\pi_{\theta_{\text{old}}}(\cdot|x,y_{<t},c)$. The base policy $\pi_{\theta}$ is then optimized to match this privileged distribution at every decoding step.
We formulate this matching objective universally using \emph{f-divergences}. For a given time step $t$, let the likelihood ratio for any token $v \in \mathcal{V}$ be defined as $\tau^t_v = \frac{\pi_{\theta_\text{old}}(v|x,y_{<t},c)}{\pi_{\theta}(v|x,y_{<t})}$. To prevent extreme ratios from destabilizing training, we clip the likelihood ratio to a bounded range:
\begin{equation*}
    \tilde{\tau}^t_v = \operatorname{clip}\!\left(\tau^t_v,\; \epsilon_{\min},\; \epsilon_{\max}\right)
\end{equation*}
The unified self-distillation objective then minimizes the expected f-divergence across all tokens $t\in[1,T]$:
\begin{equation}
\label{eq:sd}
    \mathcal{L}_{\text{SD}}(\theta) = \mathbb{E}_{x, y \sim \pi_{\theta}, f} \left[ \sum_{t=1}^{T} \sum_{v \in \mathcal{V}} \pi_\theta(v | x, y_{<t}) \cdot f(\tilde{\tau}^t_v) \right]
\end{equation}
where $f(\tau)$ is a convex function. By selecting different $f$, we can recover various divergence measures, such as forward KL, reverse KL and Jensen-Shannon Divergence (JSD), each offering unique optimization properties.
We provide a detailed discussion on the choice of $f$ and its implications in Appendix~\ref{app:divergence}.
Notice that, since we sample trajectories from the base policy $\pi_{\theta}$, the self-distillation objective is inherently on-policy. Such a formulation allows the model to recover forgotten behaviors after midtraining~\citep{shenfeld2026selfdistillationenablescontinuallearning,chen2025retainingdoingroleonpolicy,shenfeld2025rlsrazoronlinereinforcement,lai2026reinforcementfinetuningnaturallymitigates} and prevent from exposure bias~\citep{lu2025onpolicydistillation,agarwal2024policy}.

\section{Method}\label{sec:method}
\begin{figure*}
    \centering
    \includegraphics[width=0.95\linewidth]{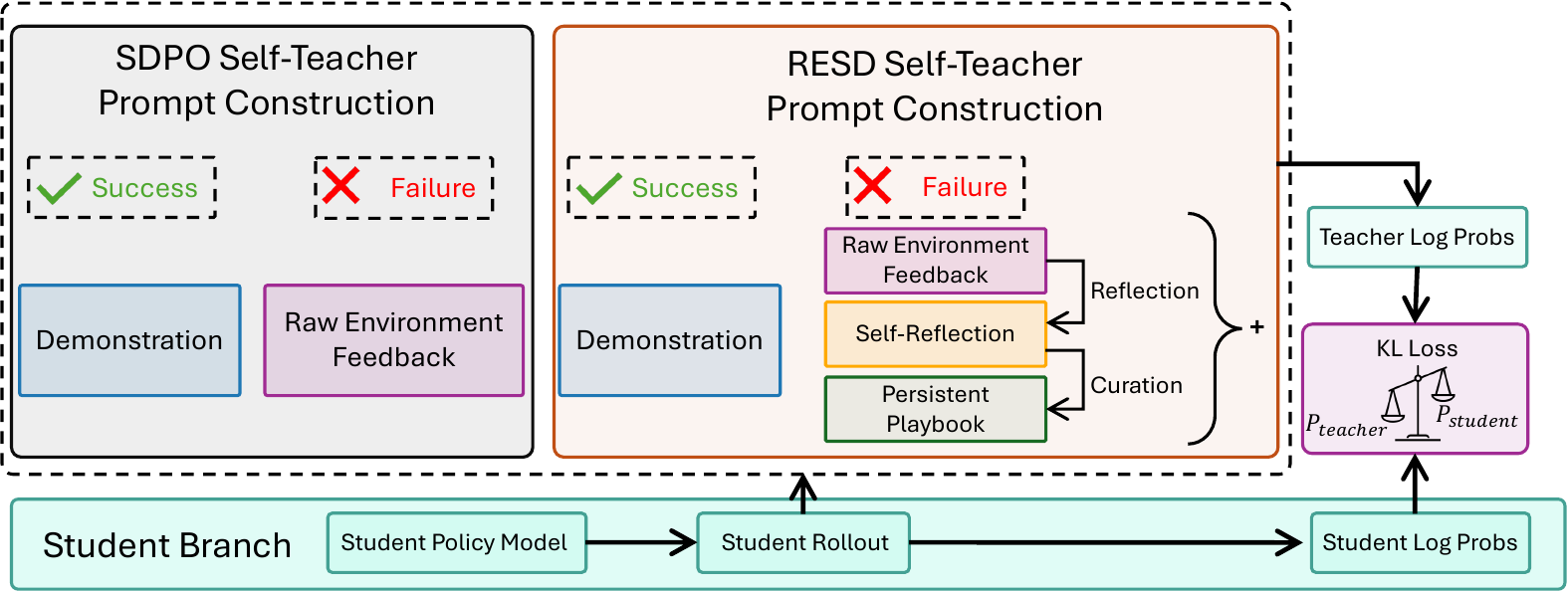}
        \caption{Overview of the \method{} framework. The student generates a rollout and receives environment feedback. On failure, a local self-reflection diagnoses the error and a global playbook curation step distills reusable lessons into a persistent playbook. The enriched context (reflection, curated playbook, and any cached solutions) is fed to the teacher prompt, whose output distribution provides token-level supervision to the student via a KL loss.}
        % \textcolor{red}{change this plot to highlight teacher prompt construction section and remove unnecessary details}}
    \label{fig:workflow}
    \vspace{-1.5em}
\end{figure*}
Our work builds on the finding that existing self-distillation methods rely heavily on successful demonstrations, and their supervision degrades when successful rollouts are scarce. We propose Reflection-Enhanced Self-Distillation (\method{}), which enriches the teacher's context with retrospective feedback from failures and a persistent memory of past experiences, enabling corrective supervision even without successful demonstrations. We first motivate the problem by analyzing SDPO's failure mode under sparse rewards (\S\ref{subsec:sdpo_breakdown}), then detail the retrospective reflection and playbook curation mechanism (\S\ref{subsec:reflection_playbook}), and describe how these components are integrated into a memory-augmented self-distillation framework (\S\ref{subsec:memory_augmented_sd}).

\subsection{Self-Distillation Relies on Successful Peer Demonstrations}\label{subsec:sdpo_breakdown}

The privileged context of the SDPO self-teacher comprises two complementary signals: environment feedback on the current trajectory and successful peer demonstrations sampled within the same rollout batch~\citep{hubotter2026reinforcement}. To disentangle their contributions, we conduct an ablation that varies the rollout batch size $N$ while keeping all other hyperparameters fixed. When $N{=}1$, each prompt group samples a single rollout, eliminating the possibility of peer demonstrations entirely and leaving environment feedback as the sole supervisory signal. As shown in Figure~\ref{fig:success_ablation}, SDPO's performance degrades substantially in this regime, indicating that the teacher struggles to translate failure feedback alone into effective corrective distributions.
This failure mode becomes particularly severe in rare-success regime. If the per-rollout success rate is small, most rollout groups will contain only failures and provide no positive counterfactual demonstration, forcing the teacher to construct token-level supervision from raw failure feedback alone which will result in less performance improvements through training.

\subsection{From Passive Feedback Exposure to Active Feedback Understanding}\label{subsec:reflection_playbook}
\begin{wrapfigure}{r}{0.485\textwidth}
    \centering
    % \vspace{-1em}
    \includegraphics[width=0.48\textwidth]{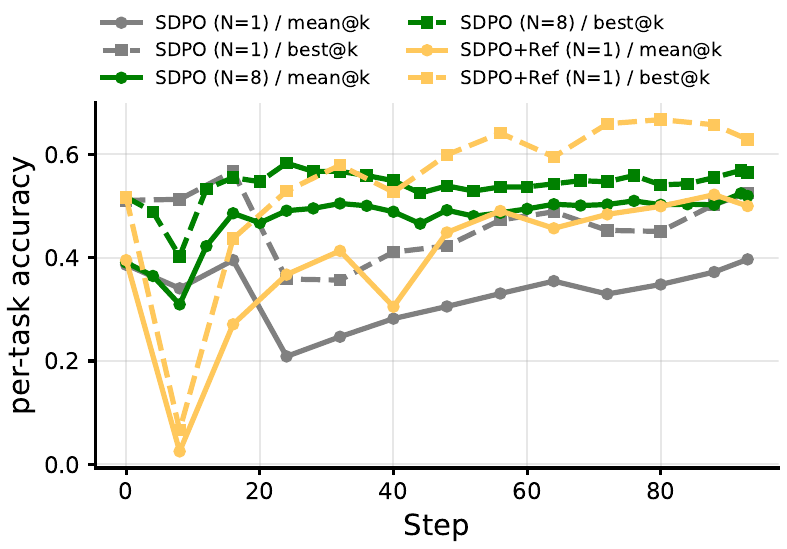}
    \caption{Per-task accuracy on \textsc{Finer} under varying rollout batch sizes. SDPO ($N{=}1$) degrades without peer demonstrations, while adding reflection and playbook curation (SDPO+Ref, $N{=}1$) recovers and surpasses the $N{=}8$ baseline.}
    \label{fig:success_ablation}
    \vspace{-1.5em}
\end{wrapfigure}
The degradation observed in the $N{=}1$ setting suggests that environment feedback alone is not sufficient to support effective self-distillation. This is somewhat surprising, as the teacher is still given privileged information about the student's failed trajectory, and in principle should be able to use this information to produce a more corrective distribution. \emph{We argue that the issue lies not in the absence of feedback but in how feedback is represented and used.} Existing self-distillation methods often expose the teacher to environment feedback as additional context, but leave the interpretation of that feedback entirely implicit. In rare-success regimes, this passive use of feedback becomes a bottleneck, raw failure signals are often diagnostic rather than instructional, indicating that a trajectory failed without explaining which decision caused the failure.

\begin{wrapfigure}{r}{0.48\textwidth}
    \centering
    \vspace{-1em}
    \begin{subfigure}{0.23\textwidth}
        \centering
        \includegraphics[width=\linewidth]{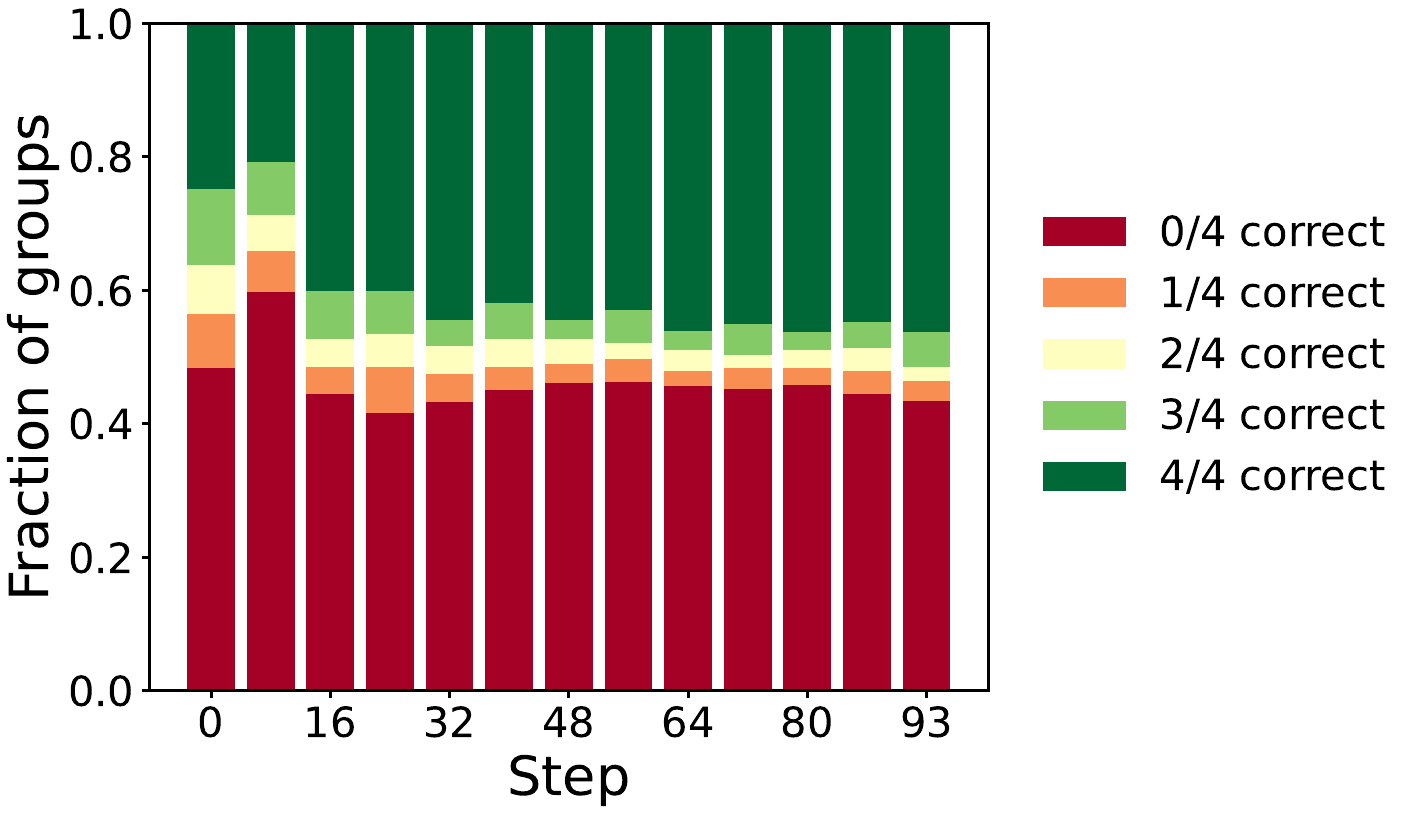}
        \caption{SDPO (N=8).}
        \label{fig:resd_top}
    \end{subfigure}
    \hfill
    \begin{subfigure}{0.23\textwidth}
        \centering
        \includegraphics[width=\linewidth]{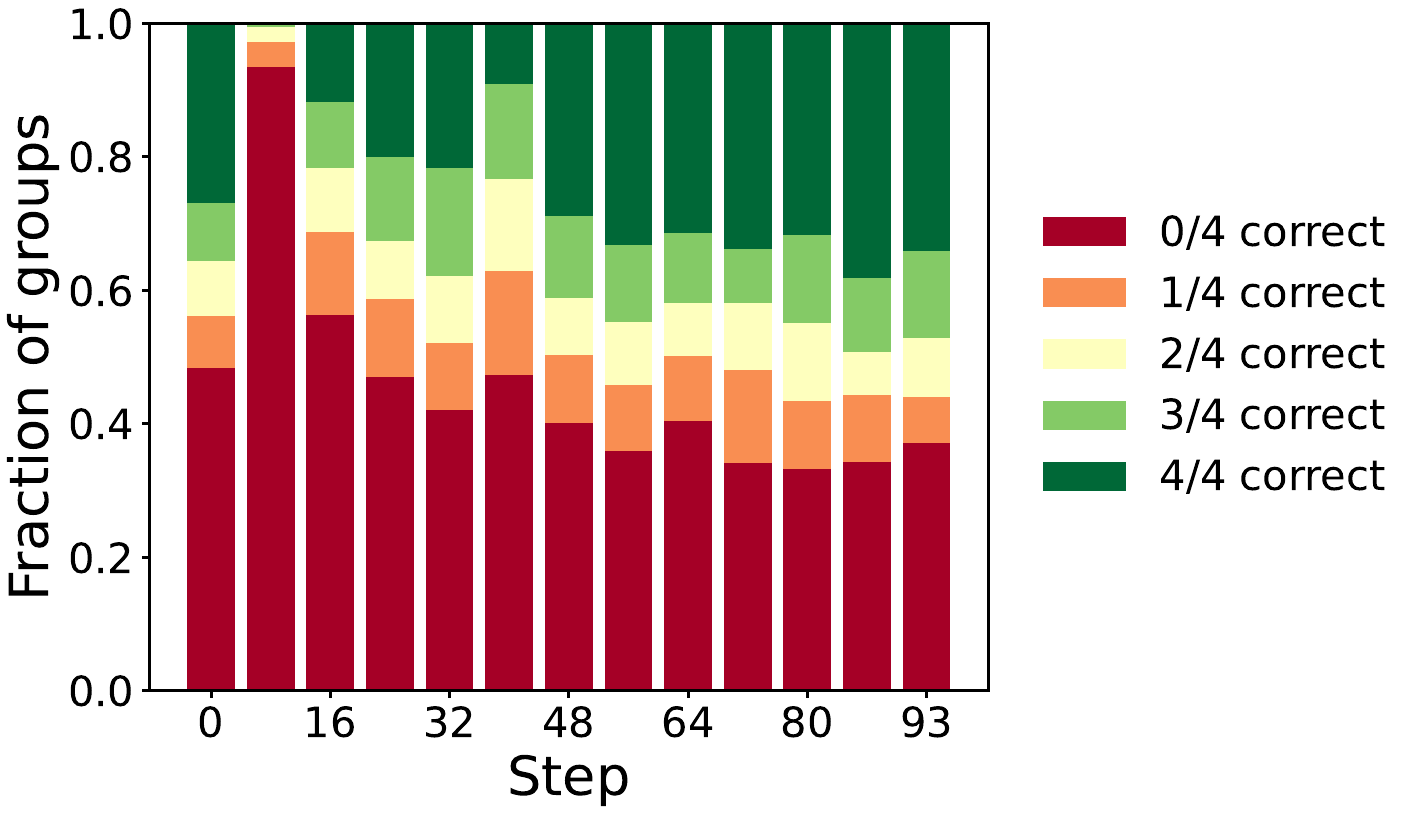}
        \caption{SDPO+Ref (N=1).}
        \label{fig:resd_bottom}
    \end{subfigure}
    \caption{Per-prompt mean accuracy distribution on \textsc{Finer} over training steps. All-wrong cases' proportion decreases better for SDPO+Ref.}
    \label{fig:distribution_analysis}
    \vspace{-1.3em}
\end{wrapfigure}

This motivates a shift from passive feedback exposure to active feedback understanding. For feedback to become useful supervision, the teacher must perform two additional operations. \textbf{(a)} It must retrospectively interpret the failed trajectory, connecting the final feedback signal to the intermediate reasoning that caused the failure. \textbf{(b)} It must preserve useful lessons across trajectories, since many failures are repeated manifestations of the same hidden rule rather than independent mistakes. In this view, raw feedback is only a starting point, and \emph{it must be transformed into causal explanations and persistent knowledge before it can reliably guide token-level distillation}. As shown by the yellow curve in Figure~\ref{fig:success_ablation}, equipping the teacher with only these two failure-derived components (without access to any successful demonstrations yet) already recovers much of the performance lost in the $N{=}1$ regime. Further analysis on the distribution of per-prompt accuracy (i.e. we sample $4$ times for each prompt and measure the proportion of correct trials) in Figure~\ref{fig:distribution_analysis} shows that the performance improvement of SDPO ($N{=}8$) mostly comes from increasing the number of completely correct prompts, while SDPO+Ref ($N{=}1$) significantly reduces the number of completely incorrect prompts, indicating that reflections help the model learn from failures.

\subsection{Reflection-Enhanced Self-Distillation (\method{})}\label{subsec:memory_augmented_sd}

Building on the above observation, we introduce Reflection-Enhanced Self-Distillation (\method{}), a framework that operationalizes active feedback understanding within the self-distillation loop.
\method{} maintains two forms of persistent context: a playbook $\mathcal{P}_t$, following the broader idea from~\citep{zhang2025agentic}, that stores reusable lessons distilled from previous failures, and an optional solution buffer $\mathcal{B}$ that caches successful trajectories when they are available.
At each training step, \method{} first updates this context using the outcome of the current rollout, and then conditions the teacher on the enriched context to produce token-level supervision. This design allows failed trajectories to improve future teacher prompts, rather than being used only as one-shot raw feedback.

\noindent \textbf{Retrospective Reflection on Failures.}
% Given a failed trajectory $(x, y)$ with environment feedback $c(x,y)$ and the current playbook $\mathcal{P}_t$, the reflector generates a structured reflection:
% \begin{equation}
%     r = \textsc{Reflect}(x, y, c, \mathcal{P}_t;\, \theta)
% \end{equation}
% where the student model $\pi_\theta$ is prompted with a diagnostic instruction that includes the original problem $x$, the failed response $y$, environment feedback $c$, and the current playbook $\mathcal{P}_t$. The reflection $r$ serves a dual purpose: it produces a natural-language analysis of the failure, and it assigns a tag $l_j \in \{\text{helpful}, \text{harmful}, \text{neutral}\}$ to each existing playbook entry $p_j \in \mathcal{P}_t$, indicating whether that entry contributed to or detracted from the failed reasoning. These tags accumulate across training steps, providing a usage signal for downstream playbook maintenance.
For a failed trajectory $(x, y)$ with environment feedback $c(x,y)$, \method{} invokes the student model in a diagnostic mode to generate a retrospective reflection. Formally, given the current playbook $\mathcal{P}_t$, the reflector outputs
\[
r,\{\ell_{t,j}\}_{j=1}^{|\mathcal{P}_t|}
=
\textsc{Reflect}(x, y, c, \mathcal{P}_t;\, \theta),
\]
where $r$ is a natural language explanation of the failure and $\ell_{t,j}\in\{\texttt{helpful},\texttt{harmful},\texttt{neutral}\}$ indicates whether the existing
playbook entry $p_j\in\mathcal{P}_t$ helped or misled the current attempt.
The reflection is designed to make the raw feedback more actionable by connecting it to specific reasoning steps. The entry-level tags provide a lightweight usage signal
for maintaining the playbook over time.

\noindent \textbf{Playbook Curation from Past Experiences.}
The playbook $\mathcal{P}_t = \{p_1, \dots, p_M\}$ is a persistent set of natural-language entries that encode reusable lessons. Given a reflection $r$, the curator generates new entries to add:
\[
    \mathcal{P}_{t+1} = \mathcal{P}_t \cup \textsc{Curate}(\mathcal{P}_t, r;\, \theta)
\]
where $\pi_\theta$ is prompted with the current playbook and the reflection, and produces a set of non-redundant entries that capture lessons not already present in $\mathcal{P}_t$. Each entry $p_j$ maintains accumulated counts $(h_j, d_j)$ of helpful and harmful tags received from the reflector across steps. Before each context update, a \textsc{Concise} operation prunes the playbook by removing entries where $d_j \geq h_j$ (net harmful) and, if $|\mathcal{P}_t|$ exceeds a budget $M_{\max}$, evicting the least recently tagged entries. This lifecycle ensures the playbook remains compact and converges toward entries that are repeatedly useful across trajectories. We detail the full \textsc{Concise} strategy, including trigger conditions and two removal variants, in Appendix~\ref{app:concise}.

\noindent \textbf{Interleaving Context Update with Policy Optimization.}
Algorithm~\ref{alg:resd} summarizes the full \method{} training loop. At each step $t$, the student $\pi_\theta$ generates a rollout $y \sim \pi_\theta(\cdot \mid x)$ and receives environment feedback $c(x,y)$ and reward $R(x,y)$ to judge the correctness. Before the policy update, a context update phase modifies the persistent state: \textsc{Concise} first prunes $\mathcal{P}_t$, then, if the rollout failed, \textsc{Reflect} and \textsc{Curate} extend the playbook to $\mathcal{P}_{t+1}$; if it succeeded, the solution buffer is updated $\mathcal{B}(x) \leftarrow y$ and sets
$r=\emptyset$. The teacher model is then synced with student model with an EMA update~\citep{hubotter2026reinforcement}, and conditioned on the enriched context to produce token-level supervision:
\[
    \pi_{\theta_\text{old}}(\cdot \mid x, y_{<t}, c, r, \mathcal{P}_{t+1}, \mathcal{B}(x))
\]
and the student minimizes the self-distillation objective $\mathcal{L}_\text{SD}(\theta)$ (Eq.~\ref{eq:sd}). We additionally apply a per-sample weight to enhance the gradient contribution of successful samples, computed as a function of the batch success rate (details in Appendix~\ref{app:sr_weight}). Because $\mathcal{P}$ and $\mathcal{B}$ persist across steps, the supervision available to the teacher can improve over time even when the current batch contains no successful demonstration.

\begin{algorithm}[t]
\caption{Reflection-Enhanced Self-Distillation (\method{})}
\label{alg:resd}
\small
\setlength{\algorithmicindent}{1em}
\begin{algorithmic}[1]
\REQUIRE Student policy $\pi_\theta$, teacher weights $\theta_\text{old}$, playbook $\mathcal{P} \leftarrow \emptyset$, solution buffer $\mathcal{B} \leftarrow \emptyset$
\FOR{each training step}
    \STATE \textbf{// Rollout \& Reward}
    \STATE Sample rollout $y \sim \pi_\theta(\cdot \mid x)$
    \STATE Obtain environment feedback $c(x, y)$ and reward $R(x, y)$
    \STATE \textbf{// Context Update}
    \STATE $\mathcal{P} \leftarrow \textsc{Concise}(\mathcal{P})$ \hfill $\triangleright$ prune stale/harmful entries
    \IF{$R(x,y) < \text{threshold}$}
        \STATE $r \leftarrow \textsc{Reflect}(x, y, c, \mathcal{P};\, \theta)$ \hfill $\triangleright$ generate reflection and tag playbook entries
        \STATE $\mathcal{P} \leftarrow \textsc{Curate}(\mathcal{P}, r;\, \theta)$ \hfill $\triangleright$ generate new playbook entries from reflection
    \ELSE
        \STATE $\mathcal{B}(x) \leftarrow y$ \hfill $\triangleright$ cache successful solution for future replay
    \ENDIF
    \STATE \textbf{// Enriched Teacher Prompt Construction}
    \STATE Retrieve buffered solution $\mathcal{B}(x)$ (if available)
    \STATE Construct enriched context: playbook $\mathcal{P}$, reflection $r$, previous trial $y$, feedback $c$, solution $\mathcal{B}(x)$
    \STATE \textbf{// Policy Update}
    \STATE Compute teacher distribution $\pi_{\theta_\text{old}}(\cdot \mid x, y_{<t}, c, r, \mathcal{P}, \mathcal{B}(x))$
    \STATE Update $\theta$ by minimizing $\mathcal{L}_\text{SD}(\theta)$
    \STATE $\theta_\text{old} \leftarrow \theta$ \hfill $\triangleright$ sync teacher weights
\ENDFOR
\end{algorithmic}
\end{algorithm}

\section{Experiments}
\label{sec:exp}

Our experiments are designed to address the following questions:
\begin{itemize}[leftmargin=1.5em,itemsep=2pt]
    \item Does \method{} improve over self-distillation baselines especially in rare-success regimes? (\S\ref{sec:main_results},\S\ref{sec:critical_teacher})
    \item How does \method{} compare to reward-based RL (GRPO) in terms of interaction efficiency? (\S\ref{sec:grpo_comparison})
    \item How does each component of \method{} affect the performance? (\S\ref{sec:ablation})
    \item How does \method{} interpret and organize failure feedback at the instance level? (\S\ref{sec:case_study})
\end{itemize}
We additionally evaluate instruction-following preservation and training latency in Appendices~\ref{app:instruction_following} and~\ref{app:latency}.

\subsection{Experimental Setup}

\noindent \textbf{Datasets.}
We evaluate on four tasks spanning program synthesis and physical reasoning. \textsc{Manufactoria-Has} and \textsc{BouncingSim-Easy}/\textsc{Medium} are drawn from the RL-Grok benchmark~\citep{sun2025rlgrok}, while \textsc{Finer} is from~\cite{zhang2025agentic}. The RL-Grok tasks are characterized by near-zero initial success rates, making them a natural testbed for learning from failure feedback. \textsc{Finer}, by contrast, exhibits a higher initial success rate, allowing us to assess whether \method{} also benefits regimes where successful demonstrations are more accessible. Table~\ref{tab:datasets} summarizes the datasets.
We train \texttt{Qwen3-4B-Thinking-2507} for \textsc{Manufactoria-Has}, \textsc{BouncingSim-Easy}, and \textsc{Finer}, and \texttt{Qwen3-30B-A3B-Thinking-2507} for the harder \textsc{BouncingSim-Medium}.

\begin{table}[t]
\centering
\caption{Overview of evaluation tasks. ``Train'' and ``Test'' denote the number of unique problems in each split.}
\label{tab:datasets}
\scriptsize
\setlength{\tabcolsep}{10pt}
\begin{tabular}{@{}lccp{0.54\textwidth}@{}}
\toprule
Task & Train & Test & Description \\
\midrule
\mbox{\textsc{Manufactoria-Has}~\citep{sun2025rlgrok}} & 742 & 132 & Write DSL programs that check whether an input tape contains a target pattern \\
\mbox{\textsc{BouncingSim-Easy}~\citep{sun2025rlgrok}} & 640 & 100 & Write Python code to simulate 2-D multi-object bouncing dynamics (easy) \\
\mbox{\textsc{BouncingSim-Medium}~\citep{sun2025rlgrok}} & 320 & 100 & Write Python code to simulate 2-D multi-object bouncing dynamics (medium) \\
\mbox{\textsc{Finer}~\citep{zhang2025agentic}} & 1000 & 500 & Tag financial named entities in SEC filings with XBRL labels \\
\bottomrule
\end{tabular}
\vspace{-1.3em}
\end{table}

\noindent \textbf{Settings.}
We adopt an \emph{online streaming} training protocol in which the model makes a single pass over the training data and each training example is seen at most once.
For every incoming batch, the trainer executes an inner loop of up to $K=4$ update iterations on the same set of prompts.
This protocol is designed to mimic
practical self-improvement scenarios, where deployed models continuously encounter new tasks, observe task outcomes, and must update under a limited interaction budget ($N{=}1$).
Unless otherwise specified, we adopt EMA update for teacher model weights with an update rate of $0.0001$. We optimize reverse-KL divergence as the self-distillation loss~\ref{eq:sd} with a lower clip $\epsilon_{\text{min}} = 0.2$ and no upper clip, under train data batch size of $32$. We provide further implementation details in Appendix~\ref{app:implementation}.

\noindent \textbf{Metrics and Baselines.}
We report both \emph{mean@4} (m@4), the average performance over four independently sampled responses per test problem, and \emph{best@4} (b@4), the best performance among the four samples. We use two evaluation metrics: \emph{Per-Task Accuracy}, which measures the fraction of problems whose response passes all test cases (equivalent to pass@1 for a single sample, also used as reward function), and \emph{Per-Test-Case Accuracy}, which measures the average fraction of test cases passed across problems. Since different validation metrics may peak at different training steps, we select a single checkpoint per run using a rank-based rule: for each logged step, we rank its four validation metrics, sum the ranks, and report the checkpoint with the highest total rank. The same selection rule is applied uniformly to all methods.
We additionally include \textbf{SDPO+ss}, a variant that also distills on the self-success instead of skipping them (disabled as a default behavior in SDPO), providing denser teacher supervision in higher-success regimes.

\begin{table}[t]
\centering
\caption{Main results on four tasks. \emph{m@4} and \emph{b@4} denote mean@4 and best@4. Per-Task Accuracy is the fraction of problems whose response passes all test cases; Per-Test-Case Accuracy is the average fraction of test cases passed.}
\label{tab:main_results}
\resizebox{0.95\textwidth}{!}{%
\begin{tabular}{@{}l@{\hskip 2em}c@{\hskip 1em}c@{\hskip 2em}c@{\hskip 1em}c@{\hskip 2em}c@{\hskip 1em}c@{\hskip 2em}c@{\hskip 1em}c@{}}
\toprule
& \multicolumn{2}{c}{\textsc{Manufactoria-Has}} & \multicolumn{2}{c}{\textsc{BouncingSim-Easy}} & \multicolumn{2}{c}{\textsc{BouncingSim-Medium}} & \multicolumn{2}{c}{\textsc{Finer}} \\
& \multicolumn{2}{c}{\footnotesize (Qwen3-4B)} & \multicolumn{2}{c}{\footnotesize (Qwen3-4B)} & \multicolumn{2}{c}{\footnotesize (Qwen3-30B-A3B)} & \multicolumn{2}{c}{\footnotesize (Qwen3-4B)} \\
\cmidrule(lr){2-3} \cmidrule(lr){4-5} \cmidrule(lr){6-7} \cmidrule(lr){8-9}
Method & m@4 & b@4 & m@4 & b@4 & m@4 & b@4 & m@4 & b@4 \\
\midrule
\multicolumn{9}{@{}l}{\emph{Per-Task Accuracy}} \\
\midrule
Base model & 0.38 & 1.52 & 0.00 & 0.00 & 1.00 & 3.00 & 39.28 & 51.90 \\
% GRPO~\citep{} & -- & -- & -- & -- & -- & -- & -- & -- \\
SDPO~\citep{hubotter2026reinforcement} & 0.95 & 3.79 & 1.00 & 3.00 & 2.00 & 6.00 & 40.93 & 50.50 \\
SDPO+ss & 0.57 & 2.27 & 1.00 & 4.00 & 2.00 & 6.00 & 50.25 & 59.12 \\
\method{} (ours) & 35.80 & 65.91 & 4.25 & 8.00 & 7.25 & 10.00 & 53.66 & 61.12 \\
\midrule
\multicolumn{9}{@{}l}{\emph{Per-Test-Case Accuracy}} \\
\midrule
Base model & 25.31 & 60.97 & 27.85 & 33.53 & 23.72 & 34.67 & 69.65 & 79.21 \\
% GRPO~\citep{} & -- & -- & -- & -- & -- & -- & -- & -- \\
SDPO~\citep{hubotter2026reinforcement} & 37.66 & 73.55 & 28.98 & 46.20 & 24.38 & 34.47 & 71.27 & 78.21 \\
SDPO+ss & 37.97 & 71.93 & 29.68 & 47.73 & 24.93 & 35.80 & 76.15 & 82.21 \\
\method{} (ours) & 76.95 & 96.53 & 38.87 & 55.80 & 34.72 & 41.07 & 78.98 & 84.32 \\
\bottomrule
\end{tabular}%
}
\vspace{-1.1em}
\end{table}

\subsection{Main Results}
\label{sec:main_results}

% Table~\ref{tab:main_results} reports the main results across all four tasks. For each metric we report \emph{mean@4} (average accuracy over 4 i.i.d.\ samples per test problem) and \emph{best@4} (best-of-4). \emph{Per-Task Accuracy} measures the fraction of problems whose generated solution passes \emph{all} test cases, while \emph{Per-Test-Case Accuracy} averages the fraction of test cases passed across all problems. Best results on each column are shown in \textbf{bold}. Since different validation metrics may peak at different training steps, we apply a rank-based selection rule to pick a single checkpoint per run rather than cherry-picking the step that maximizes any individual metric. Concretely, for each of the four validation metrics we rank every logged step from worst to best (ties share their average rank), sum the four ranks per step, and report numbers from the step with the highest total rank. The same selection rule is applied uniformly to all methods.
% Table~\ref{tab:main_results} reports the main results across all four tasks. Full validation training curves for every run in Table~\ref{tab:main_results} are provided in Appendix~\ref{app:main_curves}.

Table~\ref{tab:main_results} reports the main results across all four tasks. \method{} achieves the best performance across all tasks and metrics. The gains are largest in the rare-success regimes. On \textsc{Manufactoria-Has}, \method{} improves per-task accuracy from $0.57/2.27$ under SDPO+ss to $35.80/65.91$ in m@4/b@4, while also improving per-test-case accuracy from $37.97/71.93$ to $76.95/96.53$. \method{} also consistently improves over SDPO and SDPO+ss on both \textsc{BouncingSim} tasks, showing that failure-derived reflection and playbook memory are useful beyond the DSL setting. On \textsc{Finer}, where successful demonstrations are more accessible, \method{} still obtains the strongest results, suggesting that structured feedback remains beneficial even outside the most sparse-success regime. Full validation training curves for every run in Table~\ref{tab:main_results} are provided in Appendix~\ref{app:main_curves}.

\subsection{Reflection and Playbook Elicit a More Corrective Teacher}
\label{sec:critical_teacher}

% \textcolor{red}{per-token loss distribution}

% \begin{wrapfigure}{r}{0.55\textwidth}
%     \centering
%     \vspace{-1em}
%     \begin{subfigure}{0.27\textwidth}
%         \centering
%         \includegraphics[width=\linewidth]{figs/manufactoria_has_loss.pdf}
%         \caption{\textsc{Manufactoria-Has}}
%         \label{fig:critical_teacher_left}
%     \end{subfigure}
%     \hfill
%     \begin{subfigure}{0.27\textwidth}
%         \centering
%         \includegraphics[width=\linewidth]{figs/bouncingsim_medium_loss.pdf}
%         \caption{\textsc{BouncingSim-Medium}}
%         \label{fig:critical_teacher_right}
%     \end{subfigure}
%     \caption{Self-distillation loss curves.}
%     \label{fig:loss_curves}
%     \vspace{-1.1em}
% \end{wrapfigure}
\begin{figure}[t]
    \centering
    \includegraphics[width=0.95\linewidth]{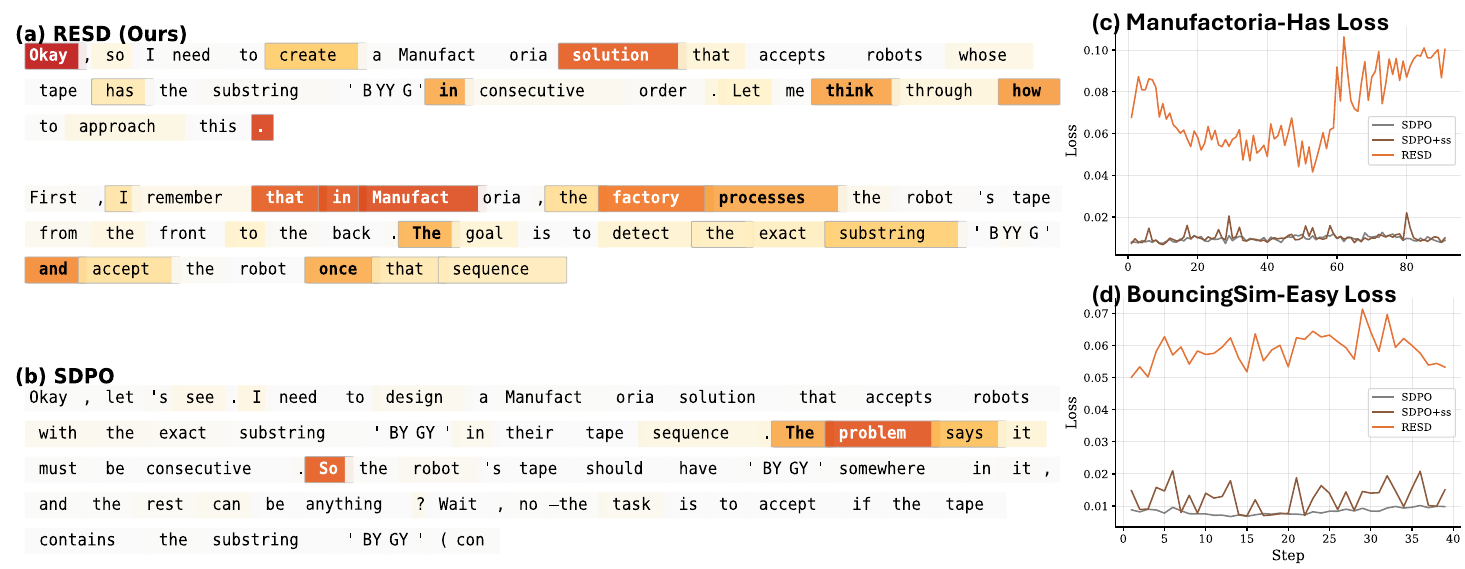}
    \caption{(a)(b) Token-level distillation loss for training step 20. Each token is shaded from white (zero loss) to dark red (high loss). (c)(d) Average per-token distillation loss across training steps for \method{} and baselines.}
    \label{fig:loss_curves}
    \vspace{-1.3em}
\end{figure}

We next examine how enriching the teacher prompt changes the self-distillation signal. Figure~\ref{fig:loss_curves} plots both per-token and average per-token distillation losses over training steps. (1) Across tasks, \method{} often yields a higher self-distillation loss than the baselines. Since this loss measures the discrepancy between the student distribution and the feedback-conditioned teacher distribution, a higher value suggests that the reflection- and playbook-augmented teacher provides a distribution that deviates more from the student's current behavior.
(2) \method{} has more loss mass focusing on reasoning decision tokens, like ``create'', ``solution'', ``processes'' and ``The goal''. These are points where the models choose what to do next or how to frame the problem, which demonstrate that the teacher is correcting the reasoning strategy.

\subsection{Comparison with GRPO}
\label{sec:grpo_comparison}

\begin{figure}[t]
    \centering
    \begin{subfigure}{0.48\textwidth}
        \centering
        \includegraphics[width=\linewidth]{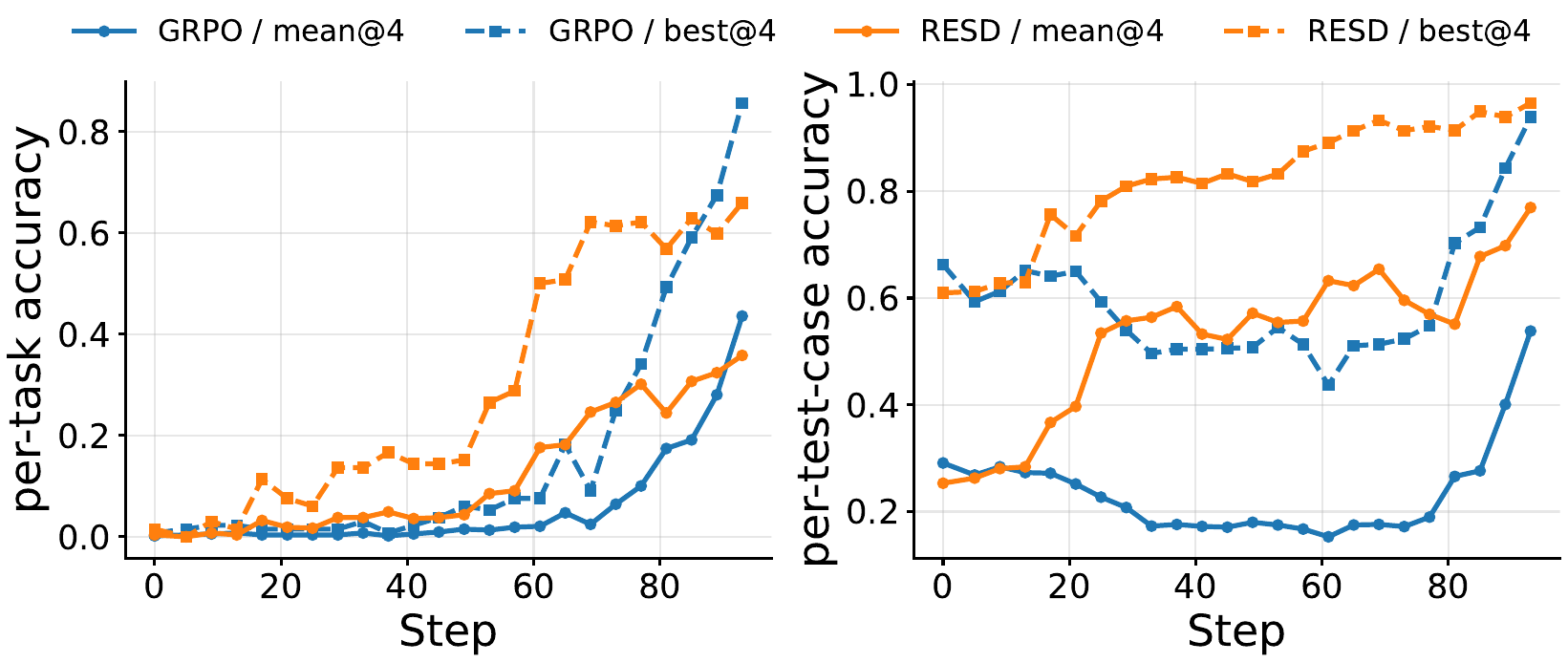}
        \caption{\textsc{Manufactoria-Has}}
        \label{fig:grpo_topleft}
    \end{subfigure}
    \hfill
    \begin{subfigure}{0.48\textwidth}
        \centering
        \includegraphics[width=\linewidth]{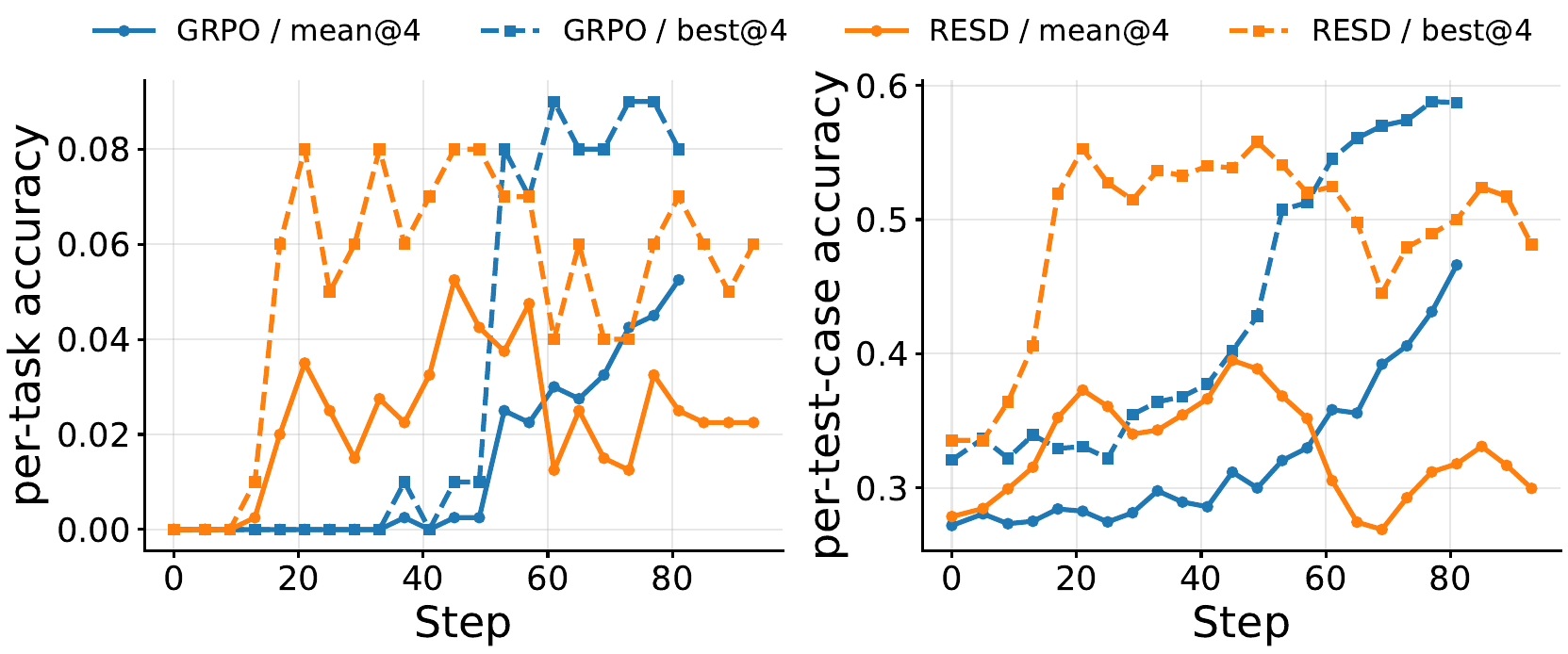}
        \caption{\textsc{BouncingSim-Easy}}
        \label{fig:grpo_topright}
    \end{subfigure}

    \vspace{-0.1em}

    \begin{subfigure}{0.48\textwidth}
        \centering
        \includegraphics[width=\linewidth]{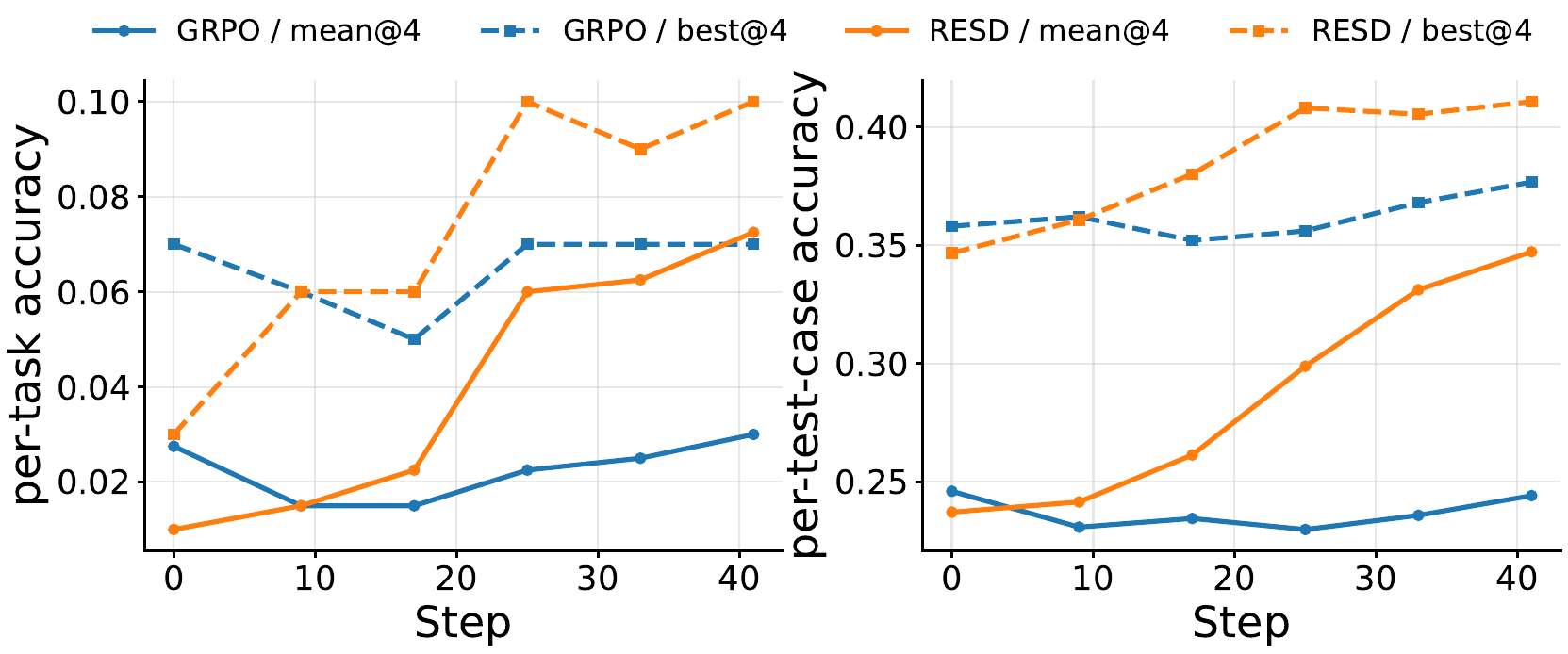}
        \caption{\textsc{BouncingSim-Medium}}
        \label{fig:grpo_bottomleft}
    \end{subfigure}
    \hfill
    \begin{subfigure}{0.48\textwidth}
        \centering
        \includegraphics[width=\linewidth]{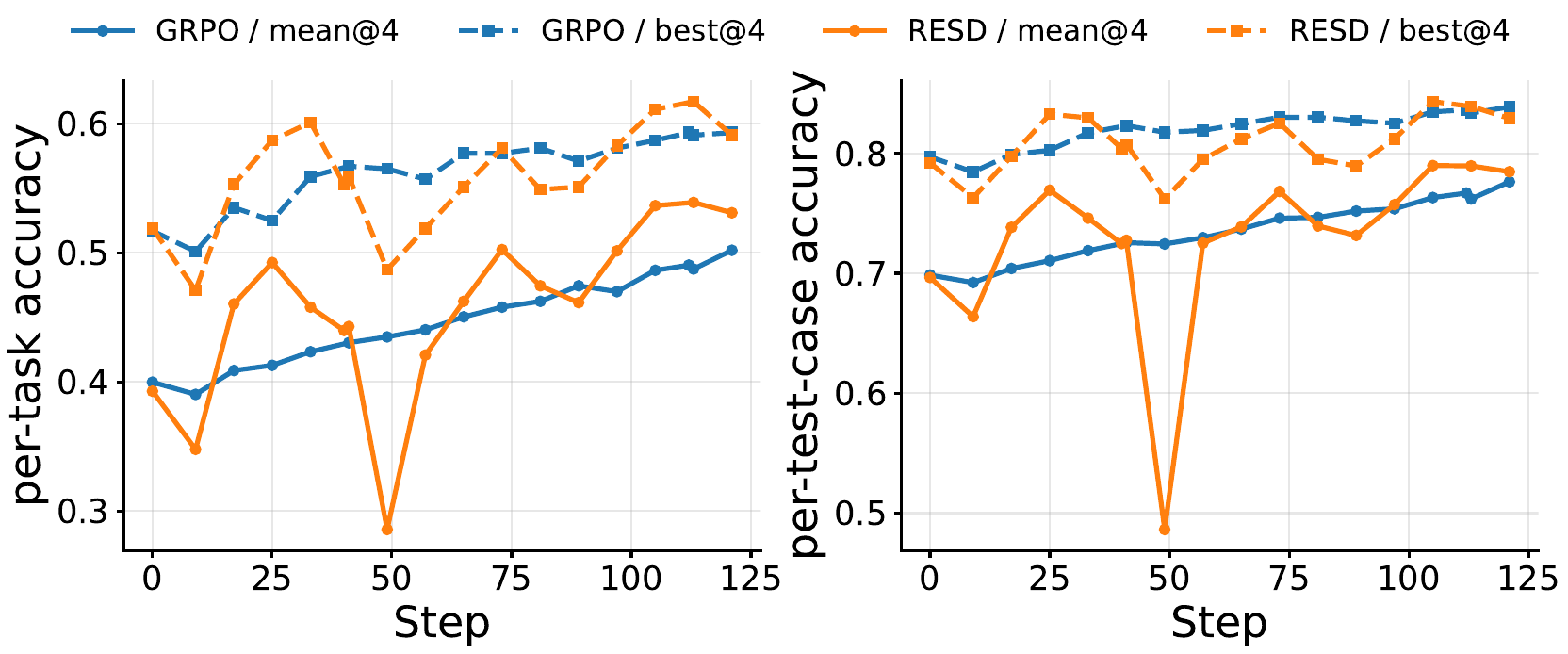}
        \caption{\textsc{Finer}}
        \label{fig:grpo_bottomright}
    \end{subfigure}
    \caption{Validation performance across training steps for \method{} and GRPO. GRPO samples $8$ rollouts per prompt, while \method{} uses a single rollout.}
    % Despite using $8\times$ fewer environment interactions per prompt, RESD improves faster in the early training stage across all four tasks, indicating stronger sample efficiency when interactions are limited.}
    \label{fig:grpo_comparison}
    \vspace{-2.1em}
\end{figure}

% \textcolor{red}{explain why there is a decrease at last or in the middle}

We compare \method{} with GRPO~\citep{shao2024deepseekmath}, a strong on-policy reinforcement learning baseline, in Figure~\ref{fig:grpo_comparison}. The key difference is the interaction pattern: GRPO uses grouped rollouts with group size $8$, requiring eight sampled trajectories and reward/feedback evaluations per prompt, whereas \method{} uses a single rollout per prompt and updates its teacher context from that trajectory. We therefore view this comparison as an interaction-efficiency comparison rather than a strict equal-rollout-budget comparison.
Despite using fewer rollouts per prompt, \method{} improves faster than GRPO in the early stage of training across the tasks in Figure~\ref{fig:grpo_comparison}. These results suggest that \method{} is well suited for sample-efficient bootstrapping under limited interaction budgets.
We note that \method{} does not always improve monotonically and the validation curves exhibit occasional dips. On \textsc{Finer}, the sharp drop around step $50$ is primarily caused by response truncation. On \textsc{BouncingSim-Easy}, the later-stage degradation after around step $60$ suggests that once the model reaches a higher-success regime, dense self-distillation from feedback-derived contexts becomes less stable than reward-based optimization.
A practical strategy is to use \method{} to quickly reach a higher-success regime, then switch to stabler reward-based optimization such as GRPO once grouped rollouts provide reliable learning signals.

\subsection{Ablation Study on Teacher Prompt Components}
\label{sec:ablation}

\begin{wraptable}{r}{0.55\textwidth}
    \centering
    \vspace{-1em}
    \caption{Ablation of teacher prompt components. We report per-test-case accuracy (m@4 / b@4).}
    \label{tab:ablation}
    \scriptsize
    \setlength{\tabcolsep}{3pt}
    \begin{tabular}{@{}l@{\hskip 1.5em}c@{\hskip 0.8em}c@{\hskip 1.5em}c@{\hskip 0.8em}c@{\hskip 1.5em}c@{\hskip 0.8em}c@{}}
    \toprule
    & \multicolumn{2}{c}{\textsc{Manuf.-Has}} & \multicolumn{2}{c}{\textsc{BSim-Easy}} & \multicolumn{2}{c}{\textsc{Finer}} \\
    \cmidrule(lr){2-3} \cmidrule(lr){4-5} \cmidrule(lr){6-7}
    Method & m@4 & b@4 & m@4 & b@4 & m@4 & b@4 \\
    \midrule
    \method{} (full) & 76.95 & 96.53 & 38.87 & 55.80 & 78.98 & 84.32 \\
    \quad w/o reflection & 51.41 & 79.26 & 36.85 & 55.33 & 77.78 & 82.26 \\
    \quad w/o playbook & 60.65 & 92.37 & 32.95 & 53.60 & 77.42 & 83.07 \\
    \quad w/o solution buffer & 60.24 & 80.10 & 31.97 & 52.40 & 78.36 & 83.42 \\
    \bottomrule
    \end{tabular}
    \vspace{-1em}
\end{wraptable}

We ablate \method{}'s enriched teacher context by removing local reflection, persistent playbook, or solution buffer. Table~\ref{tab:ablation} shows that full \method{} achieves the strongest performance across all three tasks. Removing reflection causes the largest degradation on \textsc{Manufactoria-Has} (76.95$\to$51.41 m@4), while removing the playbook or solution buffer also lowers performance, suggesting that local reflection provides useful short-term corrections while the playbook preserves reusable lessons and prevents regressions. The solution buffer further helps convert partial progress into fully correct solutions.

\subsection{Case Study: How \method{} Interprets Failure Feedback}
\label{sec:case_study}

% \textcolor{red}{explain the manufactoria task}

Table~\ref{tab:case_study_brbr} traces \method{} on a representative \textsc{Manufactoria-Has} task across four consecutive inner-loop updates, illustrating how local environment feedback is progressively organized into reusable task knowledge.
This case highlights two mechanisms that are difficult to see from aggregate accuracy alone. \emph{First,
\method{} exposes learning progress before sparse success appears}: the model fixes syntax, then
transition handling, then a more subtle edge case involving empty-tape behavior. \emph{Second, the playbook accumulates situation-specific rules}: different failure modes require different corrections (e.g., skipping non-pattern symbols vs.\ rejecting when the tape is empty), and the playbook records each rule alongside the situation it applies to. A cached successful exemplar ultimately demonstrates how these rules coexist in a complete working program. This supports the
view that failure feedback becomes useful not merely by being appended to the prompt, but by being
organized into persistent, scoped knowledge.

\begin{table}[t]
\centering
\small
\caption{
Case study on \textsc{Manufactoria-Has}: ``Accept if the tape contains BRBR.''
In this task, the model must write a finite-state machine program in a domain-specific language that processes an input tape---a sequence of colored symbols (e.g., \texttt{GRBRBRY}, \texttt{BRGBY})---one symbol at a time from left to right. Each state reads the next symbol and branches to a successor state depending on the symbol; a tape is accepted by reaching a designated accept state and rejected otherwise. Here, the target program must accept any tape containing the contiguous substring BRBR (e.g., accept \texttt{GRBRBRY}, reject \texttt{BRGBY}), requiring the model to handle arbitrary surrounding symbols, restart matching after partial mismatches, and correctly reject when the input ends mid-match.
``acc.'' denotes per-test-case accuracy. Each row corresponds to one inner-loop update step, showing how \method{} progressively diagnoses and corrects errors through reflection and playbook accumulation. Full generated programs for each step are provided in Appendix~\ref{app:case_study}.
}
\label{tab:case_study_brbr}
\begin{tabular}{@{}c c p{4.2cm} p{7.4cm}@{}}
\toprule
Step & acc. & Failure reason & Reflection / playbook update \\
\midrule
45 & 0.00
& Program references a non-existent state, causing a parse error
& Reflection extracts a syntax rule: all branch targets must reference states that are explicitly declared in the program. \\
\midrule
46 & 0.84
& Incorrectly rejects tapes whose leading symbols are not part of the pattern (e.g., rejects \texttt{GBRBR} because it starts with \texttt{G}, even though BRBR appears later)
& Reflection identifies that state \texttt{s0} routes a leading \texttt{G} to reject instead of continuing the search. Playbook records: non-pattern symbols before or between matches should be skipped, not rejected. \\
\midrule
47 & 0.96
& Overcorrection: the fix for skipping non-pattern symbols introduces an infinite loop between two states when the tape becomes empty at a non-terminal state
& Reflection identifies that non-pattern symbols should be skipped when the tape is non-empty, but the program should reject when the tape is empty at a non-terminal state. The playbook records both situation-specific rules. \\
\midrule
48 & 1.00
& --- & A successful solution is cached. \\
% & A cached successful solution from a prior correct rollout provides a complete working program that correctly handles both situations---skipping non-pattern symbols and rejecting on empty tape---yielding a fully correct solution. \\
\bottomrule
\end{tabular}
\vspace{-1.3em}
\end{table}

\section{Related Works}

\noindent \textbf{On-policy self-distillation.}
Knowledge distillation is widely used as an alternative to supervised fine-tuning (SFT) when a stronger teacher model is available. Classical approaches train a student to match the output distribution or representations of a teacher~\citep{kim2016sequence,sanh2019distilbert,team2024gemma,ko2024distillm}. However, standard distillation relies on fixed data and suffers from student--teacher mismatch, since the student is evaluated on its own generated trajectories~\citep{agarwal2023gkd,xu2024dpo,chen2024self}.
To reduce this distribution shift, recent works study on-policy distillation~\citep{lu2025onpolicydistillation,xiao2026mimo,yang2025qwen3,deepseekai2026deepseekv4}, where the student learns from teacher feedback on its own generations. This aligns the training distribution with the student policy and is closely related to online imitation learning.
Building on this idea, self-distillation uses a self-teacher that has access to privileged information to guide the current policy. The teacher may condition on environment feedback or execution signals~\citep{hubotter2026reinforcement,ye2026policy}, or leverage reference solutions and hints~\citep{zhao2026self,agarwal2024policy}. Other works explicitly study privileged information distillation in this setting~\citep{penaloza2026privileged,chen2025retainingdoingroleonpolicy}. While these methods are restricted to isolated instance-level corrections, our approach introduces cross-sample reflection to aggregate experience and provide a richer training signal.

% \paragraph{Continual learning via self-reflection and experience accumulation.}
\noindent \textbf{Continual learning via self-reflection and experience accumulation.}
Recent work explores training-free continual learning by extracting reusable strategies from an LLM's own trajectories. Models can utilize verbal feedback as episodic memory~\citep{shinn2023reflexion,madaan2023self}, self-discover reasoning structures~\citep{zhou2024self,wang2023self}, or maintain an evolving experience pool to dynamically inject refined strategies into future inference~\citep{zhang2025agentic,yoran2023answering}. Building on this trajectory-based memory paradigm, our method integrates experience accumulation directly into on-policy self-distillation.

\section{Discussion and Conclusions}
\label{sec:conclusion}
% \begin{itemize}
%     \item summary
%     \item context update \& model update
%     \item limitations of current self-distillation objective, our work is orthogonal to the design of objectives
% \end{itemize}

In this work, we address the challenge of continuous LLM self-improvement via environmental interactions, particularly in sparse-reward regimes. We introduce Reflection-Enhanced Self-Distillation (\method{}), a framework that utilizes retrospective reflections and a persistent playbook to successfully transform raw failure feedback into active, corrective supervision. Empirical evaluations across multiple complex reasoning tasks demonstrate that \method{} significantly enhances interaction efficiency. By relying on only a single rollout per prompt, it achieves rapid early-stage convergence, substantially outperforming standard self-distillation methods and GRPO.
% Crucially, our study reveals that feedback formulation is a vital design axis that is orthogonal to the self-distillation loss objective.

\noindent \textbf{\method{} as a feedback-enhancement module.}
The non-monotonic validation curves in Figure~\ref{fig:grpo_comparison} highlight an important property of on-policy self-distillation: its effectiveness depends not only on the structure and reliability of the teacher context, but also on the optimization objective. Recent studies similarly observe that self-distillation can exhibit unstable dynamics~\citep{kim2026does,li2026unifying,yang2026self}. Our results complement this line of work from a different angle. Rather than proposing a new self-distillation objective, \method{} improves the information supplied to the self-teacher by converting raw failure feedback into local reflections and persistent playbook knowledge. This explains why \method{} is especially effective in the early rare-success regime, where reward-only methods obtain little signal and standard self-distillation lacks successful demonstrations. More broadly, \method{} can be plugged into stronger self-distillation algorithms: sample routing can decide when to apply distillation~\citep{li2026unifying}, reward-grounded objectives can determine update directions~\citep{yang2026self}, and \method{} can provide the structured feedback context that makes failed trajectories useful for token-level supervision.

% \clearpage

\bibliographystyle{plain} % Or use {plain} or {unsrtnat}
\bibliography{ref} % The name of your .bib file (without the .bib extension)

%%%%%%%%%%%%%%%%%%%%%%%%%%%%%%%%%%%%%%%%%%%%%%%%%%%%%%%%%%%%

\appendix

\section{Discussion on the Self-Distillation Objective}\label{app:divergence}

As established in Section~\ref{sec:prelim}, our unified self-distillation objective minimizes the expected $f$-divergence between the teacher distribution $\pi_{\theta_\text{old}}(\cdot|x,y_{<t},c)$ and the student policy $\pi_\theta(\cdot|x,y_{<t})$. The choice of the generator function $f$ yields qualitatively different optimization landscapes. Below we discuss three canonical instantiations and their practical implications.

\paragraph{Forward Kullback-Leibler (KL) Divergence.}
Setting $f(\tau) = \tau \log \tau$ recovers the standard distillation objective. Because the loss is an expectation weighted by the student policy $\pi_\theta$ yet penalizes deviations through $\tau \log \tau$, the resulting gradient is dominated by regions where the teacher assigns substantial probability mass. This produces \emph{mean-seeking} behavior: the student is penalized heavily whenever it assigns low probability to tokens the teacher considers valid, encouraging it to cover all modes of the teacher distribution.

\paragraph{Reverse KL Divergence.}
Setting $f(\tau) = -\log \tau$ swaps the directional penalty, yielding \emph{mode-seeking} behavior. Here the student incurs large loss when it generates tokens that the teacher deems highly improbable. In complex reasoning tasks, this property is particularly desirable because it actively suppresses logical hallucinations—tokens that are plausible under the student's distribution but inconsistent with the teacher's feedback-informed reasoning.

\paragraph{Jensen-Shannon Divergence (JSD).}
Setting $f(\tau) = \frac{\tau}{2}\log \tau - \frac{\tau+1}{2}\log\!\left(\frac{\tau+1}{2}\right)$ yields a smoothed, symmetric objective bounded between $0$ and $\log 2$. By interpolating between the forward and reverse directions, JSD stabilizes training in regimes where the teacher and student distributions are nearly disjoint—precisely the situation that causes extreme gradient explosions under pure forward or reverse KL. This boundedness makes JSD a robust default when the quality gap between teacher and student is large or highly variable across training.

\section{Success-Rate Rebalancing}\label{app:sr_weight}

In the self-distillation loss, every sample in a rollout batch contributes equally by default.
When the batch success rate is very low (or very high), the gradient signal is dominated by the majority outcome, diluting the supervision from the minority.
To counteract this imbalance, we weight each sample $i$ by a scalar $w_i$ that depends on the batch success rate $s = |\{i : R(x_i, y_i) \geq \tau\}| / B$:
\begin{equation}
    w_i =
    \begin{cases}
        (1 - s)^{\alpha} & \text{if } R(x_i, y_i) \geq \tau \quad (\text{success}), \\
        s^{\beta} & \text{otherwise} \quad (\text{failure}),
    \end{cases}
\end{equation}
where $\alpha, \beta > 0$ are hyperparameters that control the strength of rebalancing, and $\tau$ is the success reward threshold.
The weights are normalized so that their batch mean equals one, $\bar{w} = \frac{1}{B}\sum_i w_i = 1$, ensuring that the effective learning rate is unaffected.
The weighted self-distillation loss becomes:
\begin{equation}
    \mathcal{L}_{\text{SD}}^{w}(\theta) = \frac{1}{B}\sum_{i=1}^{B} w_i \sum_{t=1}^{T_i} \sum_{v \in \mathcal{V}} \pi_\theta(v \mid x_i, y_{i,<t}) \cdot f\!\left(\tilde{\tau}^t_v\right).
\end{equation}
When success is rare ($s \to 0$), successful samples receive high weights while failed samples receive low weights, amplifying the distillation signal from the few successful demonstrations.
Conversely, when success is common, the corrective signal from failures is amplified.
In all experiments we set $\alpha = \beta = 1$.

\section{Training Curves for Main Results}\label{app:main_curves}

\begin{figure}[h]
    \centering
    \begin{subfigure}{0.48\textwidth}
        \centering
        \includegraphics[width=\linewidth]{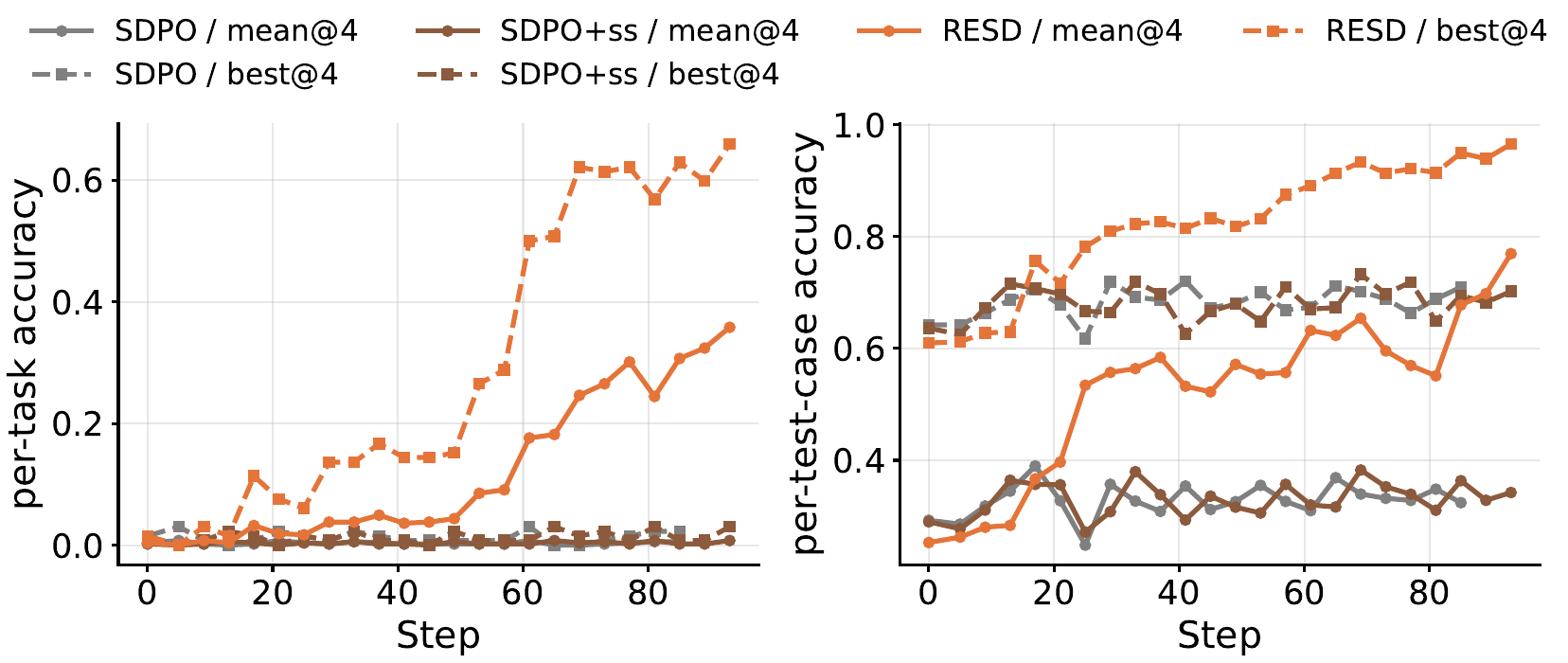}
        \caption{\textsc{Manufactoria-Has}.}
        \label{fig:main_curves_manufactoria}
    \end{subfigure}
    \hfill
    \begin{subfigure}{0.48\textwidth}
        \centering
        \includegraphics[width=\linewidth]{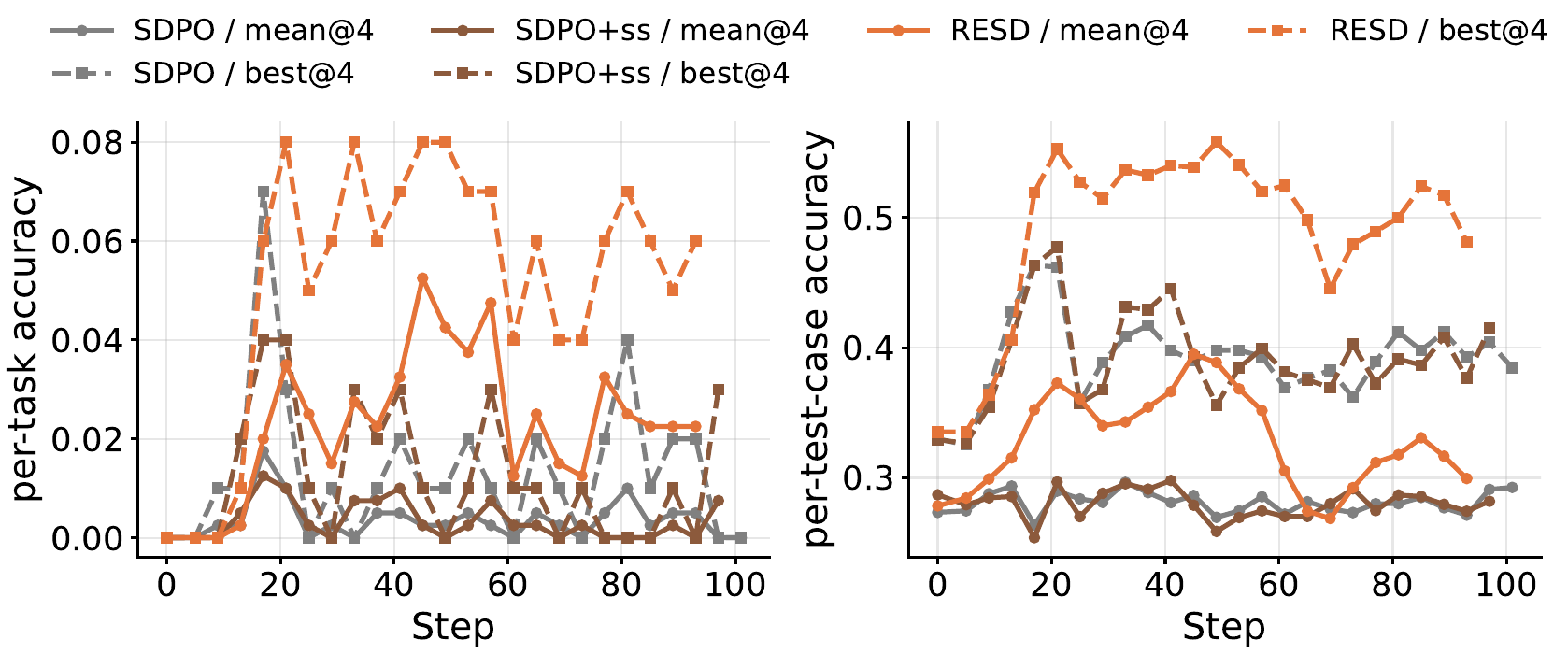}
        \caption{\textsc{BouncingSim-Easy}.}
        \label{fig:main_curves_bsim_easy}
    \end{subfigure}

    \vspace{0.5em}

    \begin{subfigure}{0.48\textwidth}
        \centering
        \includegraphics[width=\linewidth]{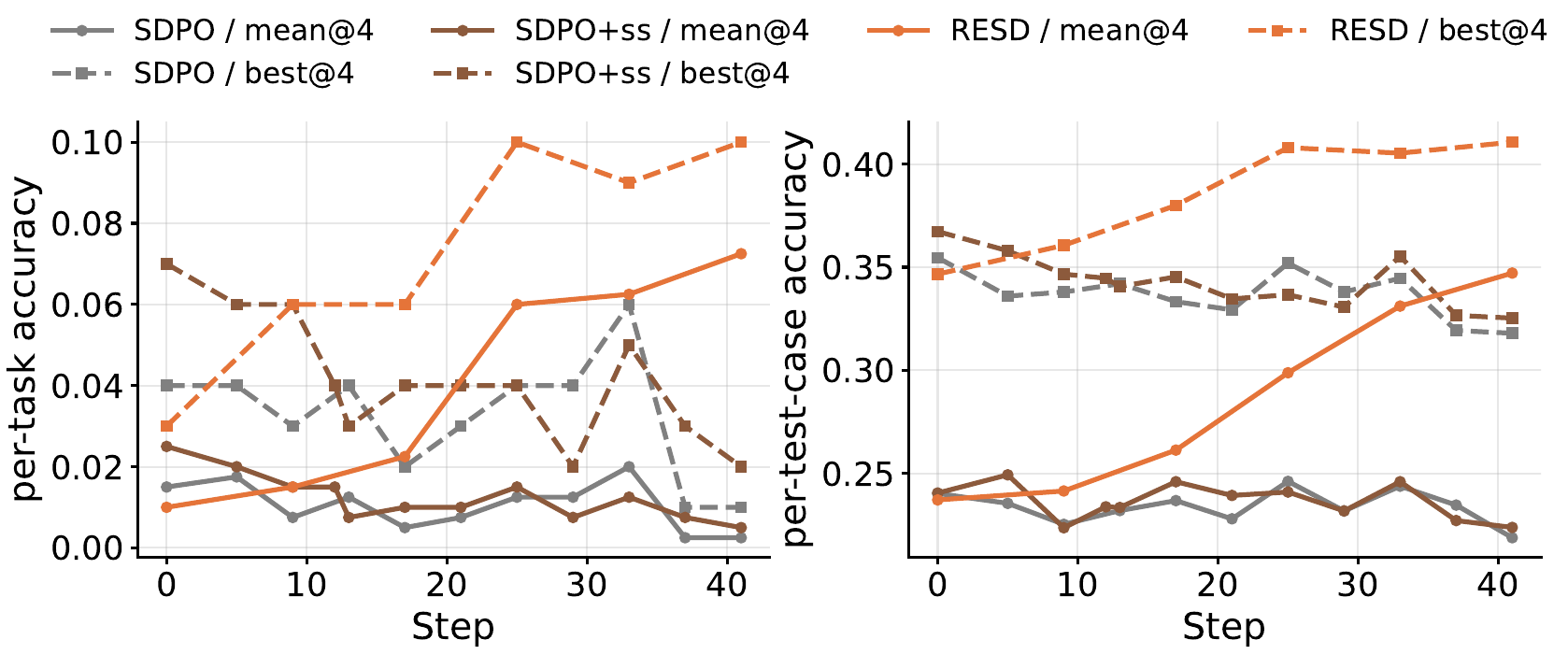}
        \caption{\textsc{BouncingSim-Medium}.}
        \label{fig:main_curves_bsim_medium}
    \end{subfigure}
    \hfill
    \begin{subfigure}{0.48\textwidth}
        \centering
        \includegraphics[width=\linewidth]{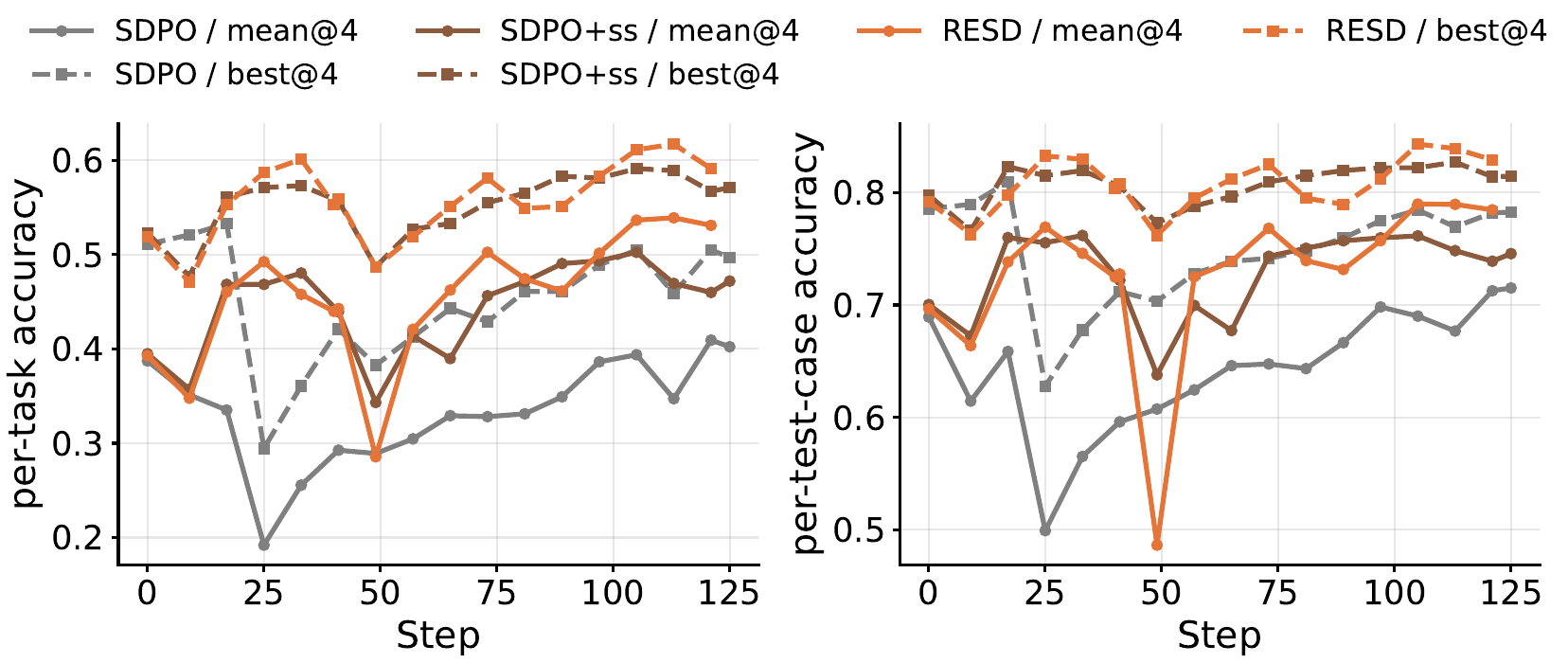}
        \caption{\textsc{Finer}.}
        \label{fig:main_curves_finer}
    \end{subfigure}
    \caption{Validation accuracy over training steps for the runs reported in Table~\ref{tab:main_results}.}
    \label{fig:main_curves}
\end{figure}

Figure~\ref{fig:main_curves} shows the validation training curves of the runs summarized in Table~\ref{tab:main_results}. Each panel plots validation accuracy over training steps for one task, comparing the baselines and \method{} under the same rank-based checkpoint-selection rule described in Section~\ref{sec:exp}.

\section{Playbook Concise Strategy}\label{app:concise}

The \textsc{Concise} operation in Section~\ref{subsec:memory_augmented_sd} maintains the playbook $\mathcal{P}$ as a compact set of reusable lessons over the course of training. We describe two concrete instantiations used in our experiments and the trigger conditions under which they are applied.

\paragraph{Bullet categorization.}
At the time of each \textsc{Concise} call, every entry $p_j \in \mathcal{P}$ is classified based on its tag history. Let $h_j$ and $d_j$ denote the accumulated helpful and harmful counts assigned by the reflector across training steps (Section~\ref{subsec:memory_augmented_sd}). We partition the playbook as
\begin{itemize}
    \item \textbf{Unused:} $h_j + d_j = 0$ (never tagged),
    \item \textbf{Harmful:} $d_j \geq h_j$ and $d_j > 0$ (net harmful),
    \item \textbf{Helpful:} otherwise.
\end{itemize}
In addition, we record the most recent step $\tau_j$ at which each entry was tagged and use $\text{step} - \tau_j$ as a staleness measure.

\paragraph{Concise methods.}
Given a budget $M_{\max}$ on the playbook size, we consider two removal strategies:
\begin{itemize}
    \item \emph{Prioritized.} Remove all unused and harmful entries. If $|\mathcal{P}|$ still exceeds $M_{\max}$, randomly drop helpful entries until the cap is met.
    \item \emph{Staleness.} Remove all harmful entries and all \emph{stale unused} entries (unused for more than one context-update step). If $|\mathcal{P}|$ still exceeds $M_{\max}$, additionally evict the longest-unused remaining entries in order of increasing $\tau_j$ until the cap is met.
\end{itemize}
Both strategies guarantee that persistently harmful entries are removed; they differ in how they treat unused entries. The \emph{staleness} variant is more patient in that unused entries are given one additional step to be tagged before becoming eligible for removal, which helps preserve newly curated lessons that may take a few reflections before receiving a usage signal.

\paragraph{Trigger conditions.}
\textsc{Concise} is invoked under two conditions: (i) a \emph{frequency trigger}, which fires every $F$ context-update steps (default $F = 4$), and (ii) a \emph{limit trigger}, which fires whenever $|\mathcal{P}| > M_{\max}$ between scheduled invocations. In the former case, the playbook is pruned down to its harmful/stale subset without enforcing the budget; in the latter, the budget $M_{\max}$ is enforced.

\paragraph{Default configuration.}
Unless otherwise noted, we use the \emph{staleness} method with $M_{\max}$ and $F$ set to the values in Appendix~\ref{app:main_curves} (training details). This is the configuration used for all \method{} runs in Table~\ref{tab:main_results}.

% Dataset details moved to Section~\ref{sec:exp}.
\label{app:datasets}

\section{Preservation of Instruction Following Capability}
\label{app:instruction_following}

\begin{table}[h]
    \centering
    \caption{IFEval accuracy after training on each task.}
    \label{tab:instruction_following}
    \scriptsize
    \setlength{\tabcolsep}{4pt}
    \begin{tabular}{@{}lcccc@{}}
    \toprule
    & \multicolumn{2}{c}{Per-Test-Case Acc.} & \multicolumn{2}{c}{Per-Task Acc.} \\
    \cmidrule(lr){2-3} \cmidrule(lr){4-5}
    Checkpoint & m@4 & b@4 & m@4 & b@4 \\
    \midrule
    Base model & 87.62 & 91.33 & 82.50 & 87.00 \\
    After \textsc{Manuf.-Has} & 87.96 & 91.33 & 83.75 & 87.00 \\
    After \textsc{BSim-Easy} & 87.83 & 92.33 & 83.25 & 88.00 \\
    After \textsc{Finer} & 86.88 & 90.33 & 80.50 & 86.00 \\
    \bottomrule
    \end{tabular}
\end{table}

A common challenge in continual learning is catastrophic forgetting, where improvements on specific reasoning tasks may degrade the model's foundational instruction-following capabilities. To verify that \method{} avoids this trade-off, we evaluate the models on the IFEval benchmark~\citep{zhou2023instruction} before and after training. As shown in Table~\ref{tab:instruction_following}, \method{} successfully preserves the model's general capabilities across all evaluation settings. For instance, after training on the \textsc{Manufactoria-Has} task, the per-task mean@4 accuracy actually increases slightly from 82.50\% to 83.75\%, while the per-test-case mean@4 accuracy remains stable at around 87--88\%.

\section{Analysis on Training Latency}
\label{app:latency}

\begin{figure}[h]
    \centering
    \includegraphics[width=0.45\textwidth]{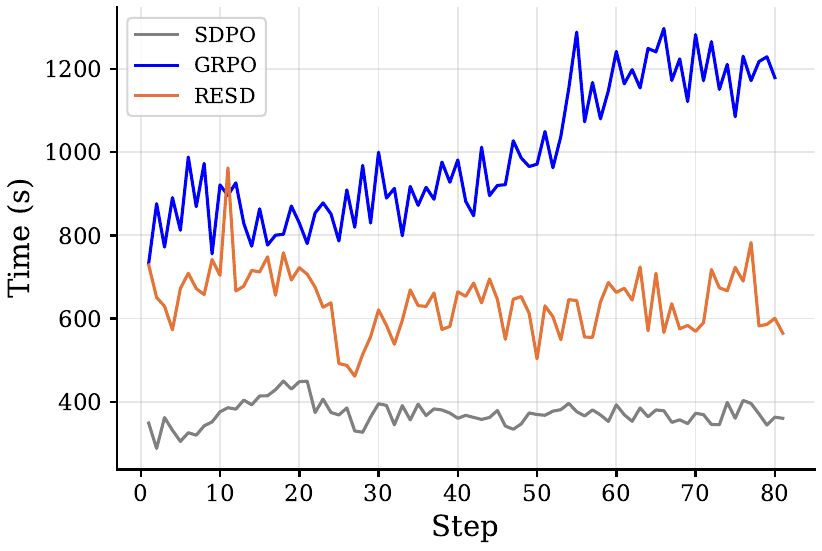}
    \caption{Per-step training latency for \textsc{BouncingSim-Easy}.}
    \label{fig:latency}
\end{figure}

While \method{} significantly improves interaction efficiency by requiring only a single rollout per prompt, the generation of retrospective reflections and playbook curation introduces additional inference steps. To understand the latency implications of these components, we analyze the per-step training latency of \method{} compared to SDPO and GRPO baselines.
As illustrated in Figure~\ref{fig:latency}, standard SDPO exhibits the lowest latency about $400$s per-step, because it only computes the teacher distribution and KL loss for a single trajectory. \method{} introduces a moderate and stable computational overhead due to the diagnostic inference required for the reflection and curation operations. However, \method{} remains substantially faster than GRPO. Because GRPO relies on group-relative optimization, it must repeatedly sample, execute, and evaluate a large batch of trajectories, causing its latency to climb above $1200$s per step.

\section{Implementation Details}
\label{app:implementation}

\paragraph{Infrastructure.}
All experiments are conducted on a single node with 8 NVIDIA H200 GPUs. Each experiment completes in approximately one day of wall-clock time. We use vLLM for rollout generation with tensor parallelism across 4 GPUs, and FSDP for distributed training. Rollout generation uses temperature $1.0$ with top-$p$ sampling at $0.95$. All models have thinking mode enabled.

\paragraph{Training.}
We train for a single epoch over the training data (online streaming). Each batch undergoes $K=4$ inner-loop update iterations. The batch size is $32$ and we use a single rollout per prompt ($N{=}1$) for all \method{} experiments. We apply a learning rate warmup of 10 steps. For validation, we sample $4$ responses per test problem to compute mean@4 and best@4 metrics.

\paragraph{Self-distillation.}
The self-distillation loss is computed over the top-$100$ tokens in the student's vocabulary at each position. Success-rate rebalancing (Appendix~\ref{app:sr_weight}) is applied on \textsc{BouncingSim-Easy} and \textsc{Finer}.

\paragraph{Context updater (\method{}-specific).}
The playbook is maintained globally (shared across all training examples). Playbook entries are tagged as helpful or harmful during reflection on both failed and successful samples. The solution buffer caches successful trajectories for future teacher prompt construction.

\paragraph{Per-task hyperparameters.}
Unless otherwise noted, we use EMA update rate $0.0001$, lower clip $\epsilon_{\min}{=}0.2$, and no upper clip. Task-specific deviations are as follows:
\begin{itemize}[nosep,leftmargin=*]
    \item \textsc{Manufactoria-Has}: reverse KL ($\alpha{=}1.0$), upper clip $\epsilon_{\max}{=}5$, max playbook entries $M_{\max}{=}200$, prioritized concise method, max prompt/response length 49152/20480 tokens.
    \item \textsc{BouncingSim-Easy}: JSD ($\alpha{=}0.5$), max playbook entries $M_{\max}{=}120$, staleness-based concise method, success-rate reweighting enabled, max prompt/response length 58368/25600 tokens.
    \item \textsc{BouncingSim-Medium}: reverse KL ($\alpha{=}1.0$), EMA update rate $0.01$, max playbook entries $M_{\max}{=}150$, prioritized concise method, max prompt/response length 58368/25600 tokens.
    \item \textsc{Finer}: JSD ($\alpha{=}0.5$), max playbook entries $M_{\max}{=}120$, staleness-based concise method, success-rate reweighting enabled, max prompt/response length 49152/20480 tokens.
\end{itemize}

\paragraph{GRPO baseline.}
For GRPO, we use a rollout group size of $8$ per prompt with the same learning rate ($1\text{e-}6$), batch size ($32$), and inner-loop steps ($K{=}4$). GRPO does not use self-distillation, context updating, or EMA.

\section{Case Study: Full Program Outputs}\label{app:case_study}

This appendix provides the complete DSL programs generated at each inner-loop step for the case study in Section~\ref{sec:case_study} (``Accept if the tape contains BRBR''). Each listing corresponds to one row of Table~\ref{tab:case_study_brbr}. We highlight the buggy lines with inline comments.

\paragraph{Step 45 (acc.\ = 0.00): Parse error from undefined node reference.}
The program references \texttt{state0} in a \texttt{PULLER\_YG} node, but \texttt{state0} is never declared as a node, causing a parse error.

\noindent\begin{minipage}{\linewidth}
\begin{lstlisting}[basicstyle=\ttfamily\scriptsize, frame=single, xleftmargin=1em]
START start:
    NEXT node_rb

PULLER_RB node_rb:
    [B] node_b1
    [R] node_rb
    [EMPTY] node_yg

PULLER_YG node_yg:
    [Y] node_rb
    [G] node_rb
    [EMPTY] state0       # BUG: 'state0' is never declared

PULLER_RB node_b1:
    [R] node_b1r1
    [B] node_b1
    [EMPTY] node_yg

...

END end
\end{lstlisting}
\end{minipage}

\paragraph{Step 46 (acc.\ = 0.84): Correct pattern matching logic, but rejects non-R/B characters.}
The state machine correctly implements substring matching transitions for B and R. However, all \texttt{[EMPTY]} branches route to \texttt{NONE}, which causes immediate rejection when a non-R/B character (Y or G) is encountered---e.g., input \texttt{GGYYYBBRBR} is rejected at state \texttt{s0} because the leading \texttt{G} triggers \texttt{[EMPTY] NONE}.

\noindent\begin{minipage}{\linewidth}
\begin{lstlisting}[basicstyle=\ttfamily\scriptsize, frame=single, xleftmargin=1em]
START start:
    NEXT s0

PULLER_RB s0:
    [B] s1
    [R] s0
    [EMPTY] NONE          # BUG: rejects Y/G characters

PULLER_RB s1:
    [R] s2
    [B] s1
    [EMPTY] NONE          # BUG: same issue

PULLER_RB s2:
    [B] s3
    [R] s0
    [EMPTY] NONE          # BUG: same issue

PULLER_RB s3:
    [R] end
    [B] s1
    [EMPTY] NONE          # BUG: same issue

END end
\end{lstlisting}
\end{minipage}

\paragraph{Step 47 (acc.\ = 0.96): Overcorrection introduces an infinite loop.}
The model adds a \texttt{PULLER\_YG} node to handle non-R/B characters, fixing the step 46 bug. However, when the tape becomes empty at a non-terminal state, \texttt{PULLER\_RB} routes to \texttt{state0\_yg}, whose \texttt{[EMPTY]} routes to \texttt{NONE}---but if there are remaining characters that alternate between the two node types, they bounce back and forth infinitely.

\noindent\begin{minipage}{\linewidth}
\begin{lstlisting}[basicstyle=\ttfamily\scriptsize, frame=single, xleftmargin=1em]
START start:
    NEXT state0

PULLER_RB state0:
    [B] state1
    [R] state0
    [EMPTY] state0_yg     # Routes non-R/B to PULLER_YG

PULLER_RB state1:
    [R] state2
    [B] state0
    [EMPTY] state0_yg

PULLER_RB state2:
    [B] state3
    [R] state0
    [EMPTY] state0_yg

PULLER_RB state3:
    [R] end
    [B] state0
    [EMPTY] state0_yg

PULLER_YG state0_yg:
    [Y] state0
    [G] state0
    [EMPTY] NONE          # BUG: causes infinite loop when tape
                          # is empty between state0 and state0_yg

END end
\end{lstlisting}
\end{minipage}

\paragraph{Step 48 (acc.\ = 1.00): Correct solution.}
The final program correctly handles all cases: non-pattern symbols are skipped via dedicated \texttt{PULLER\_YG} nodes, and empty-tape transitions at non-terminal states correctly reject. A cached successful solution from a prior rollout provided the complete exemplar that guided this output.

\noindent\begin{minipage}{\linewidth}
\begin{lstlisting}[basicstyle=\ttfamily\scriptsize, frame=single, xleftmargin=1em]
START start:
    NEXT s0

PULLER_RB s0:
    [B] s1
    [R] s0
    [EMPTY] s0_yg

PULLER_YG s0_yg:
    [Y] s0
    [G] s0
    [EMPTY] NONE

PULLER_RB s1:
    [R] s2
    [B] s1
    [EMPTY] s1_yg

PULLER_YG s1_yg:
    [Y] s1
    [G] s1
    [EMPTY] NONE

PULLER_RB s2:
    [B] s3
    [R] s0
    [EMPTY] s2_yg

PULLER_YG s2_yg:
    [Y] s2
    [G] s2
    [EMPTY] NONE

PULLER_RB s3:
    [R] end
    [B] s1
    [EMPTY] s3_yg

PULLER_YG s3_yg:
    [Y] s3
    [G] s3
    [EMPTY] NONE

END end
\end{lstlisting}
\end{minipage}

%%%%%%%%%%%%%%%%%%%%%%%%%%%%%%%%%%%%%%%%%%%%%%%%%%%%%%%%%%%%

\newpage

\end{document}